\PassOptionsToPackage{numbers,sort&compress}{natbib}
\documentclass{article}

\usepackage{arxiv}
\usepackage{float}
\usepackage[utf8]{inputenc} 
\usepackage[T1]{fontenc}    
\usepackage{hyperref}       
\usepackage{url}            
\usepackage{booktabs}       
\usepackage{amsfonts}       
\usepackage{nicefrac}       
\usepackage{microtype}      
\usepackage{lipsum}		
\usepackage{graphicx}
\usepackage{natbib}
\usepackage{doi}
\usepackage{amsmath}

\title{Learning more physically realistic dynamics in machine-learning based weather forecasting with latent-space constraints}

\author{
Hang Fan$^{1,2,*}$, Yi Xiao$^{3,5}$, Yongquan Qu$^{1,2}$, Juan Nathaniel$^{1,2}$,\\[2pt]
\textbf{Fenghua Ling$^{3}$, Ben Fei$^{3,4,*}$, Lei Bai$^{3,*}$, Pierre Gentine$^{1,2}$}\\[6pt]
\small $^{1}$Department of Earth and Environmental Engineering, Columbia University, New York, NY, USA.\\
\small $^{2}$Learning the Earth with Artificial Intelligence and Physics (LEAP) Center, Columbia University, New York, NY, USA.\\
\small $^{3}$Shanghai Artificial Intelligence Laboratory, Shanghai, China.\\
\small $^{4}$The Chinese University of Hong Kong, Hong Kong, China.\\
\small $^{5}$Department of Computer Science and Technology, Tsinghua University, Beijing, China.\\[4pt]
\small $^\ast$Corresponding author. Email: bailei@pjlab.org.cn; benfei@cuhk.edu.hk
}

\date{}

\renewcommand{\shorttitle}{Learning Physically Realistic Atmospheric Dynamics with Latent-Space Constraints}

\hypersetup{
pdftitle={A template for the arxiv style},
pdfsubject={q-bio.NC, q-bio.QM},
pdfauthor={David S.~Hippocampus, Elias D.~Striatum},
pdfkeywords={First keyword, Second keyword, More},
}

\begin{document}

\maketitle

\begin{abstract}
Data-driven machine learning (ML) models are reshaping weather forecasting and have shown the potential to accelerate and surpass traditional physics-based approaches, leading to a second revolution in the field after data assimilation.
However, most ML forecast models are trained with weighted variable-wise losses on rollout forecasts that neglect cross-variable and spatial error covariance induced by physical coupling, often yielding overly smooth and physically unrealistic long-range forecasts.
To address this, we reformulate model training as a four-dimensional variational data assimilation (4DVar) problem that treats reanalysis data as imperfect observations. This enables the loss function to incorporate cross-variable error covariance structures that capture multivariate dependencies and their associated errors.
In practice, we approximate this objective by computing the loss in an autoencoder-learned latent space of global atmospheric states.
By encoding complex nonlinear couplings among atmospheric variables, this representation allows the high-dimensional, complex error covariance matrix in model space to be approximated as nearly diagonal in latent space, substantially simplifying implementation.
We show that rollout training with latent-space constraints improves long-term forecast skill, while better preserving fine-scale structures and physical realism than the widely used model-space loss.
Finally, we extend this framework to accommodate heterogeneous data sources, enabling the forecast model to be trained jointly on reanalysis and multi-source observations within a unified theoretical formulation.
\end{abstract}

\newcommand\blfootnote[1]{%
  \begingroup
  \renewcommand\thefootnote{}%
  \footnotetext[0]{#1}%
  \endgroup
}
\blfootnote{$^\ast$Corresponding author. Email: hf2526@columbia.edu, benfei@cuhk.edu.hk, bailei@pjlab.org.cn}


\section{Introduction}
Numerical Weather Prediction (NWP) has long served as the cornerstone of modern meteorology~\cite{charney1950numerical}, supporting vital applications such as disaster response, agricultural planning, and energy management~\cite{kalnayAtmosphericModelingData2002}. Over the past decades, substantial improvements in model resolution, physical parameterizations, and data assimilation (DA) have led to a so-called quiet revolution in NWP~\cite{bauerQuietRevolutionNumerical2015}, albeit at the cost of increasingly demanding computational resources. More recently, data-driven machine learning (ML) forecast models have demonstrated competitive—and in some cases superior—performance to traditional physics-based approaches across various aspects of both deterministic~\cite{biAccurateMediumrangeGlobal2023,lamLearningSkillfulMediumrange2023,chenFuXiCascadeMachine2023,langAIFSECMWFsDatadriven2024,chenOperationalMediumrangeDeterministic2025} and probabilistic (ensemble) forecasts~\cite{kochkovNeuralGeneralCirculation2024,zhongFuXiENSMachineLearning2025,liGenerativeEmulationWeather2024,lang2026aifs,aletSkillfulJointProbabilistic2025}, while offering orders-of-magnitude gains in computational efficiency. Despite growing enthusiasm for ML-based forecasting as the future of NWP, several fundamental limitations remain to be addressed.

A central challenge is that physical realism is not guaranteed in current ML forecast models, yet it is critical to their reliability, generalizability, and long-range forecasting capability.
We argue that, for deterministic forecast models (DFMs), this limitation stems from the widespread use of weighted variable-wise loss functions, which neglects multiscale structure of the atmosphere and the intrinsic coupling across variables.
When combined with autoregressive rollout, such training objectives often yield long-range forecasts that are numerically accurate yet overly smooth and physically unrealistic~\cite{biAccurateMediumrangeGlobal2023,lamLearningSkillfulMediumrange2023,chenFuXiCascadeMachine2023,langAIFSECMWFsDatadriven2024,chenOperationalMediumrangeDeterministic2025}.
Training probabilistic forecast models, such as with the Continuous Ranked Probability Score (CRPS)~\cite{zhongFuXiENSMachineLearning2025,aletSkillfulJointProbabilistic2025,lang2026aifs} or diffusion‑based frameworks, can somehow alleviate this spatial blurring~\cite{kochkovNeuralGeneralCirculation2024,liGenerativeEmulationWeather2024,lang2026aifs}, but they still do not ensure physical consistency across variables. 

In practice, explicitly enforcing multivariable consistency in ML-based forecast models remains highly challenging.
The atmosphere comprises many interacting dynamical modes and constraints spanning a wide range of spatial and temporal scales, and their couplings are often strongly nonlinear~\cite{kalnayAtmosphericModelingData2002}. 
Further, these nonlinear couplings evolve with the synoptic flow (known as flow dependency)~\cite{kalnayAtmosphericModelingData2002,bannisterReviewForecastError2008}, making them difficult to represent explicitly and accurately.
Most current attempts to enforce physical consistency explicitly in ML-based weather models have been largely empirical and hand-designed, i.e., relying on conservation-correction modules~\cite{shaImprovingAIWeather2025a}, weak hydrostatic-balance constraints in the loss~\cite{subramaniamImposingFundamentalDynamical2025}, or spatially filtered loss functions~\cite{kochkovNeuralGeneralCirculation2024}. 
Although effective, these empirical strategies remain insufficient to account for the diverse and complex physical consistency in the atmosphere.

A parallel limitation is the lack of a unified framework for learning atmospheric dynamics from heterogeneous data sources of varying quality.
Most ML-based forecast models~\cite{biAccurateMediumrangeGlobal2023,lamLearningSkillfulMediumrange2023,chenFuXiCascadeMachine2023,kochkovNeuralGeneralCirculation2024,langAIFSECMWFsDatadriven2024,chenOperationalMediumrangeDeterministic2025} are trained exclusively on reanalysis products, most commonly ERA5~\cite{hersbachERA5GlobalReanalysis2020}, despite their known imperfections for some variables~\cite{tangHaveSatellitePrecipitation2020, laversEvaluationERA5Precipitation2022} and in regions with limited observational coverage.
Others have recently shifted entirely to observation-only training paradigms to avoid using reanalysis data~\cite{vandal2025global,alexeGraphDOPSkilfulDatadriven2024}, assuming that the available observations are sufficient to constrain the entire dynamics of the system~\cite{falascaNeuralModelsMultiscale2025}.
However, from a DA perspective, any informative data can, in principle, contribute to improved posterior estimation, provided that its uncertainty is properly quantified~\cite{kalnayAtmosphericModelingData2002}.
This concept was recently demonstrated in ~\cite{yuvalNeuralGeneralCirculation2026}, which trained a forecast model using ERA5 together with a satellite precipitation product and achieved improved precipitation forecast skill. 
However, this approach requires extensive empirical tuning of the relative weights assigned to different data sources in the loss function, making it cumbersome in practice.
Therefore, a more principled framework that jointly leverages observations and reanalysis to train ML-based forecast models is needed to guide the development of future ML forecast models.

In this study, we reformulate the training of DFMs with a rollout strategy as a form of four-dimensional (4D) variational data assimilation (4DVar)~\cite{le1986variational,zupanskiRegionalFourDimensionalVariational1993,courtierStrategyOperationalImplementation1994,zupanskiGeneralWeakConstraint1997a} problem grounded in Bayesian theory. In this framework, reanalysis data should be treated as imperfect observations, and their error covariance matrix, $\mathbf{A}$, can therefore be used to constrain the multivariate consistency.
However, explicitly and accurately representing $\mathbf{A}$ is infeasible in practice, as its dimensionality typically exceeds $10^{12}$.
To address this, we compute the loss not in the original model space but in a latent space of the atmosphere, defined by an autoencoder (AE)~\cite{hintonReducingDimensionalityData2006a} trained on ERA5~\cite{hersbachERA5GlobalReanalysis2020}.
As demonstrated in our previous work~\cite{fanPhysicallyConsistentGlobal2026}, this latent representation effectively captures complex multivariate dependencies and exhibits approximate mutual decorrelation across latent variables, allowing the complex model-space $\mathbf{A}$ to be represented in an approximately diagonal form in the latent space.
We show that training within a latent-space mean squared error (MSE) loss enables effective rollouts, substantially improving long-range forecast skill while preserving fine-scale structures and ensuring multivariate consistency.
Finally, motivated by this 4DVar perspective, we propose a more general framework that facilitates the use of heterogeneous data sources, including both observations and reanalysis, for DFM training.

\section{A New Perspective on Training DFMs}
Existing DFMs learn the parameters of an ML model to best replicate atmospheric dynamics by fitting historical reanalysis trajectories. 
This goal is conceptually analogous to 4DVar, an operational DA method that updates prior information with time-sequenced observations to obtain the maximum a posteriori (MAP) estimate of the variables governing the evolution of a dynamical system, including, but not limited to, the initial state and model parameters~\cite{le1986variational,zupanskiRegionalFourDimensionalVariational1993,courtierStrategyOperationalImplementation1994,zupanskiGeneralWeakConstraint1997a}.
Motivated by this similarity, we recast the rollout training of DFMs as a 4DVar problem and, from this perspective, clarify the assumption underlying the widely used model-space MSE loss and its limitation.

\subsection{Interpreting DFM Training as a Special Case of 4DVar}
Assuming that (i) the system evolution is fully determined by the initial conditions \( \boldsymbol{x} \) and model parameters \( \boldsymbol{\theta} \), and (ii) errors in the prior model state \( \boldsymbol{x_b} \) (known as the background field), prior model parameters \( \boldsymbol{\theta_b} \), and observations \( \boldsymbol{y} \) are mutually independent and follow zero-mean Gaussian distributions, the 4DVar problem can be formulated as minimizing the following cost function:

\begin{equation}
J(\boldsymbol{x}, \boldsymbol{\theta}) =
\frac{1}{2} \| \boldsymbol{x} - \boldsymbol{x}_b \|^2_{\mathbf{B}^{-1}}
 + \frac{1}{2} \| \boldsymbol{\theta} - \boldsymbol{\theta}_b \|^2_{\boldsymbol{\Theta}^{-1}}
+ \frac{1}{2} \sum_{i=1}^{T} \| \mathbf{y}_i - \mathcal{H}(\mathcal{M}_{0 \rightarrow i}(\boldsymbol{x}, \boldsymbol{\theta})) \|^2_{\mathbf{R}_i^{-1}},
\label{eq:4dvar_full}
\end{equation}
where \( \mathcal{M}_{0 \rightarrow i} \) denotes the forecast operator that advances \( \boldsymbol{x} \) from time \( 0 \) to time \( i \) given \( \boldsymbol{\theta} \), \( \mathcal{H} \) denotes the observation operator that maps the model state to the observation space, and \( T \) denotes the length of the assimilation window. 
The matrices \( \mathbf{B} \), \( \boldsymbol{\Theta} \), and \( \mathbf{R}_i \) represent the error covariance of \( \boldsymbol{x_b} \), \( \boldsymbol{\theta_b} \), and \( \boldsymbol{y}_i \), respectively. The derivation of Eq.~\ref{eq:4dvar_full} is provided in the Supporting Information.

In essence, Eq.~\ref{eq:4dvar_full} defines a covariance-weighted objective that penalizes deviations of the estimated state and model parameters from their priors (first two terms) and mismatches between the model-predicted and observed trajectories in observation space (third term). 
In contrast, the DFM training optimizes only the model parameters by minimizing the mismatch between model-predicted and reanalysis trajectories, which can therefore be viewed as a special case of 4DVar under the following simplifications:
\begin{enumerate}
    \item \textbf{Perfect initial state assumption}: Assuming that the prior model state \( \boldsymbol{x}_b \), typically taken from a reanalysis dataset, is perfect and error-free during training, the first term is removed and the optimization no longer targets \( \boldsymbol{x} \).
    \item \textbf{No explicit prior on model parameters}: The parameters of ML forecast models are typically initialized randomly before training, implying that no explicit prior information is imposed on them. Accordingly, the corresponding covariance \( \boldsymbol{\Theta} \) is taken to be infinitely large, eliminating the weight of the second term.
    \item \textbf{Reanalysis as imperfect state observations}: Treating reanalysis data with the same dimensionality as the model state as the only observations, the observation operator \( \mathcal{H} \) can be omitted, and \( \mathbf{R}_i \) is replaced by the error covariance matrix \( \mathbf{A}_i \) for the reanalysis \( \boldsymbol{x}_a \) at time \( i \).
\end{enumerate}
This formulation yields a MAP estimate of the model parameters by minimizing:
\begin{equation}
J(\boldsymbol{\theta}) = \sum_{i=1}^{T} \left\| \boldsymbol{x}_{a,i} - \mathcal{M}_{0 \rightarrow i}(\boldsymbol{x}_{a,0}, \boldsymbol{\theta}) \right\|^2_{\mathbf{A}_i^{-1}},
\label{eq:theta_loss}
\end{equation}
which provides a Bayesian interpretation of DFM training. Here, \( T \) represents the number of rollout steps used in training. \( \mathbf{A}_i \) characterizes the multivariate coupling structure of reanalysis errors, which shapes the direction of forecast-error optimization.

\subsection{The Assumption Behind the Model-space MSE Loss}
\begin{figure*}[ht]
    \centering
    \includegraphics[width=1.0\linewidth]{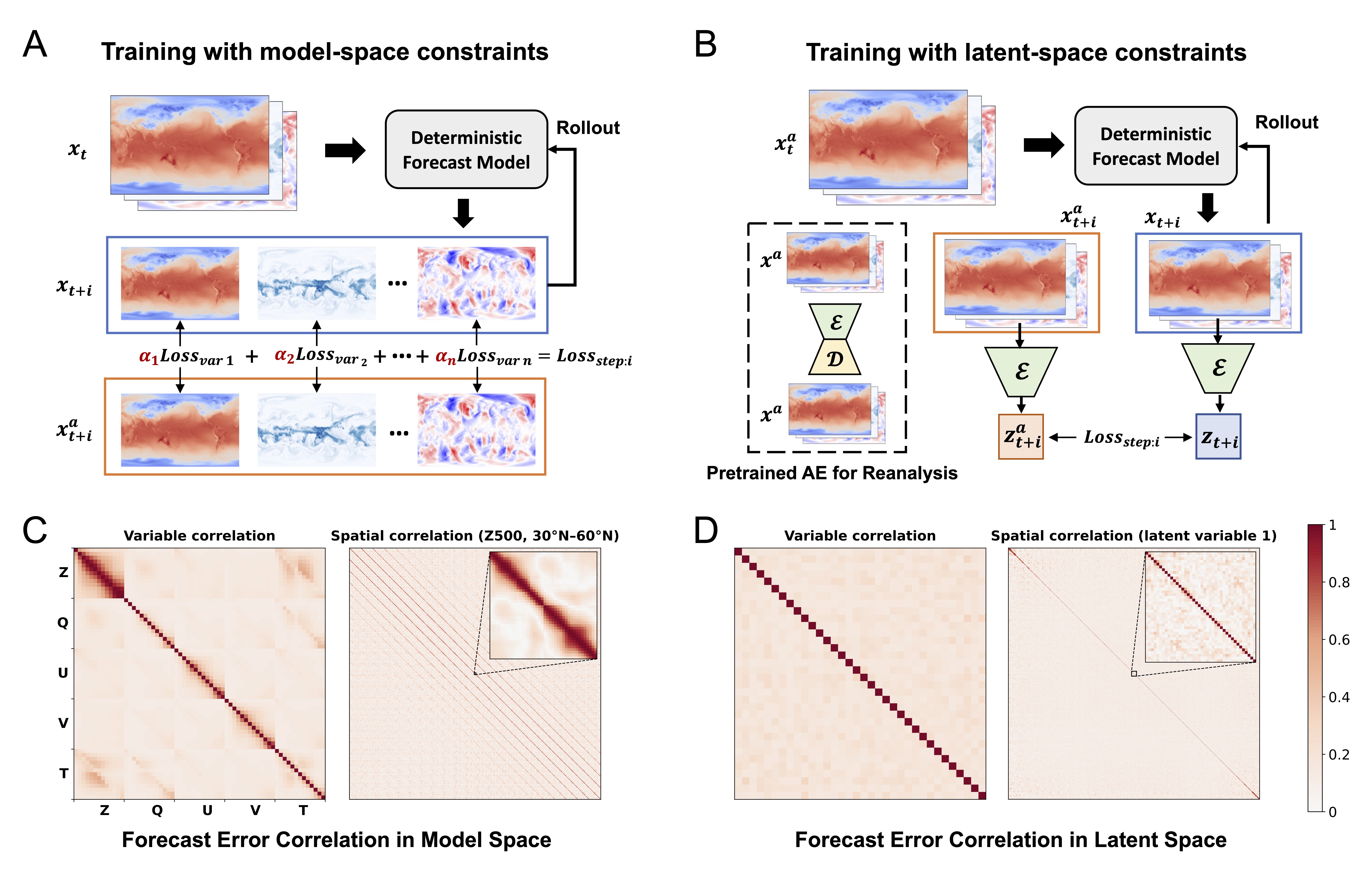}
    \caption{
    Training ML-based deterministic weather forecast models with model-space versus latent-space constraints.
    (A and B) Schematics of training DFMs in model space (A) and latent space (B).
    The superscript $a$ on model states $\boldsymbol{x}$ and latent states $\boldsymbol{z}$ indicates reanalysis data, while the subscript $i$ denotes the $i$‑th step during rollout training. 
    $\mathcal{E}$ and $\mathcal{D}$ denote the encoder and decoder of the pretrained autoencoder for reanalysis. 
    (C) Mean absolute forecast-error correlations in model space for a DFM trained without rollout. Statistics are computed from forecasts initialized daily at 0000 UTC throughout 2017 and evaluated against ERA5. Variable correlations are calculated independently at each grid point, and their absolute values are then averaged over space. Since the high dimensionality of model space, only the spatial correlation of 500-hPa geopotential height (Z500) errors over the midlatitudes is shown. (D) Same as C, but with forecast errors computed in latent space before the correlation statistics are evaluated.
    }
    \label{fig:fig1}
\end{figure*}

The weighted model-space MSE loss widely used in DFM training can be written as
\begin{equation}
J(\boldsymbol{\theta}) = \sum_{i=1}^{T} \sum_{j=1}^{n} w_{j,i} \left( \boldsymbol{x}_{a,i}^{(j)} - \mathcal{M}_{0 \rightarrow i}^{(j)}(\boldsymbol{x}_{a,0}, \boldsymbol{\theta}) \right)^2,
\label{eq:weighted_mse}
\end{equation}
where \( n \) is the number of variables, and \( w_{j,i} \) denotes the weight associated with the \( j \)-th variable at time step \( i \). 
In practice, these weights are typically determined manually through trial and error.
Comparing Eq.~\ref{eq:weighted_mse} with Eq.~\ref{eq:theta_loss} reveals that the model-space MSE loss implicitly assumes $\mathbf{A}_i^{-1}=\mathrm{diag}(w_{1,i},\ldots,w_{n,i})$, neglecting the temporal variation of reanalysis errors as well as their cross-variable covariances. 
Fig.~\ref{fig:fig1} illustrates the concept of this loss under model-space constraints, in which errors are evaluated separately for each atmospheric variable and then combined in the optimization objective as a weighted sum.

\subsection{How Reanalysis Error Covariance Shapes Physical Consistency}
In Bayesian DA methods that optimize the initial state, the covariance structure of the background error covariance matrix $\mathbf{B}$ encodes correlations among variables and thereby allows the analysis to adjust multiple variables simultaneously, which is crucial for physical consistency and realistic spatial structure~\cite{bannisterReviewForecastError2008}. 
Analogously, when optimizing a forecast model by optimizing its parameters, the covariance part of $\mathbf{A}_i$ should play a comparable role in guiding the model toward physically realistic trajectories. Below, we show how neglecting the covariance structure leads to physically unrealistic forecasts.

Let $\mathbf{D}_i = \mathrm{diag}(\mathbf{A}_i)$, the correlation matrix of the analysis error at time $i$ can then be expressed as $\mathbf{C}_i = \mathbf{D}_i^{-1/2}\mathbf{A}_i\mathbf{D}_i^{-1/2}$, with unit diagonal. Since \( \mathbf{C}_i \) is symmetric and positive definite, it admits an eigendecomposition with orthonormal principal directions:
\begin{equation}
\mathbf{C}_i = \mathbf{U}_i \boldsymbol{\Lambda}_i \mathbf{U}_i^\top,
\end{equation}
where \( \boldsymbol{\Lambda}_i = \mathrm{diag}(\lambda_{i,1},\lambda_{i,2},\ldots) \) collects the nonnegative eigenvalues of \( \mathbf{C}_i \), and \( \mathbf{U}_i = [\boldsymbol{u}_{i,1},\boldsymbol{u}_{i,2},\ldots] \) stacks the corresponding orthonormal eigenvectors.

We define \( \boldsymbol{\varepsilon}_i = \mathbf{D}_i^{-1/2}[(\boldsymbol{x}_{a,i} - \mathcal{M}_{0\rightarrow i}(\boldsymbol{x}_{a,0},\boldsymbol{\theta})] \) as the standardized form of the forecast error at time step \(i\). Expanding \( {\boldsymbol{\varepsilon}}_i \) in the eigenbasis of $\mathbf{U}_i$ gives
\begin{equation}
{\boldsymbol{\varepsilon}}_i = \sum_k \alpha_{i,k}\boldsymbol{u}_{i,k},
\end{equation}
with coefficients \( \alpha_{i,k} = \boldsymbol{u}_{i,k}^\top {\boldsymbol{\varepsilon}}_i \). Therefore, optimizing the model parameters under the full \( \mathbf{A}_i \) (i.e., Eq.~\ref{eq:theta_loss}) can be written as minimizing

\begin{equation}
J(\boldsymbol{\theta})
=
\sum_{i=1}^{T}
\boldsymbol{\varepsilon}_i^\top
\mathbf{D}_i^{1/2}\mathbf{A}_i^{-1}\mathbf{D}_i^{1/2}
\boldsymbol{\varepsilon}_i 
=
\sum_{i=1}^{T}
\boldsymbol{\varepsilon}_i^\top
\mathbf{C}_i^{-1}
\boldsymbol{\varepsilon}_i
=
\sum_{i=1}^{T}
\sum_k
\lambda_{i,k}^{-1}\alpha_{i,k}^2,
\label{eq:full_cov_eigen}
\end{equation}

with $\lambda_{i,k}^{-1}$ weighting the loss along each eigenbasis direction.
Importantly, this expression assigns smaller weights to forecast-error components projected onto the leading modes of $\mathbf{C}_i$, and larger weights to those projected onto weaker modes.
In the reanalysis error correlation statistics, the leading modes are expected to represent dominant multivariate error patterns, whereas the weaker modes represent less likely error patterns under the assumed error statistics. 
As reanalysis is typically produced by a physically constrained DA system~\cite{bannisterReviewOperationalMethods2017a}, such less likely error modes may also include patterns less consistent with physically meaningful multivariate relationships.
Consequently, the eigenvalue weighting in Eq.~\ref{eq:full_cov_eigen} tolerates dominant correlated error patterns while penalizing structural errors that are statistically less likely and physically less plausible.

When using a diagonal approximation of \( \mathbf{A}_i \), i.e., \( \mathbf{D}_i \), the cost function becomes

\begin{equation}
J(\boldsymbol{\theta})
=
\sum_{i=1}^{T}
\boldsymbol{\varepsilon}_i^\top
\mathbf{D}_i^{1/2}\mathbf{D}_i^{-1}\mathbf{D}_i^{1/2}
\boldsymbol{\varepsilon}_i 
=
\sum_{i=1}^{T}
\boldsymbol{\varepsilon}_i^\top
\boldsymbol{\varepsilon}_i
=
\sum_{i=1}^{T}
\sum_k\alpha_{i,k}^2,
\label{eq:no_eigen}
\end{equation}
which eliminates the eigenvalue-dependent weighting. This implies that error components aligned with statistically and physically meaningful structures receive the same weight as those associated with less plausible structures, potentially leading to physically unrealistic forecasts and the blurry forecast typically seen in existing DFMs.

\section{Enforcing Multivariate Consistency through Latent-Space Constraints}
To learn and predict more physically realistic dynamics in DFMs, we argue that it is necessary to incorporate the full reanalysis error covariance matrix $\mathbf{A}_i$ into the loss function. However, this is extremely challenging for three main reasons. First, similar to the background error covariance matrix $\textbf{B}$, $\textbf{A}_i$ has a dimension of over $10^{12}$ and therefore must be represented in a reduced form in practice. Second, since the true state of the atmosphere is unknown, directly estimating the error statistics of reanalysis fields is difficult. Third, like $\textbf{B}$, the structure of $\textbf{A}_i$ is flow-dependent and may evolve with synoptic conditions, which further complicates its accurate estimation.

Given the strong similarities between $\textbf{A}$ and $\textbf{B}$, the success of latent DA in simplifying $\textbf{B}$ suggests a natural route for simplifying  $\textbf{A}$ in DFM training~\cite{melinc3DVarDataAssimilation2024,zhengGeneratingUnseenNonlinear,fanNovelLatentSpace2025a,fanPhysicallyConsistentGlobal2026,melincUnifiedNeuralBackgroundError2026,fanAccurateEfficientHybridEnsemble2026}. 
Latent DA formulates its cost function in an AE-learned latent space that compresses the high-dimensional model state. 
As the latent space tends to encode multivariate dependencies in an approximately orthogonal manner to enable efficient compression~\cite{fanNovelLatentSpace2025a, fanPhysicallyConsistentGlobal2026}, the corresponding error covariance is often near-diagonal, enabling a diagonal approximation that simplifies its estimation.
In addition, several studies have shown that the AE decoder behaves linearly along directions of atmospheric variability~\cite{melinc3DVarDataAssimilation2024,fanNovelLatentSpace2025a,fanPhysicallyConsistentGlobal2026}. This property allows the latent-space variational cost to inherit a convexity structure similar to that of its model-space counterpart, thereby enabling closely related approximate solutions for DA~\cite{fanPhysicallyConsistentGlobal2026}.

Similarly, when the AE reconstruction error is negligible, and the decoder mapping is nearly affine over the optimization trajectory, the model-space DFM training loss (Eq.~\ref{eq:theta_loss}) can be approximately reformulated in latent space as (see Supporting Information for proof):
\begin{equation}
J(\boldsymbol{\theta})
\approx \sum_{i=1}^{T} \left\| \mathcal{E}(\boldsymbol{x}_{a,i})
- \mathcal{E}\!\left(\mathcal{M}_{0 \rightarrow i}(\boldsymbol{x}_{a,0}, \boldsymbol{\theta})\right)
\right\|^2_{\mathbf{A}_{z,i}^{-1}},
\label{eq:latent_theta_loss}
\end{equation}
where $\mathcal{E}$ denotes the AE encoder, and \( \mathbf{A}_z = \mathbb{E}\left[(\mathcal{E}(\boldsymbol{x}_a) - \mathcal{E}(\boldsymbol{x}_t))(\mathcal{E}(\boldsymbol{x}_a) - \mathcal{E}(\boldsymbol{x}_t))^\top\right] \) denotes the reanalysis error covariance matrix in latent space. 
Furthermore, if $\mathbf{A}_z$ is approximately diagonal, Eq.~\ref{eq:latent_theta_loss} can be further reduced to:
\begin{equation}
J(\boldsymbol{\theta}) \approx \sum_{i=1}^{T} \sum_{j=1}^{m} k_{j,i} \left( \mathcal{E}^{(j)}(\boldsymbol{x}_{a,i}) - \mathcal{E}^{(j)}(\mathcal{M}_{0 \rightarrow i}(\boldsymbol{x}_{a,0}, \boldsymbol{\theta})) \right)^2,
\label{eq:weighted_latent_loss}
\end{equation}
where \( m \) is the number of latent variables, and \( k_{j,i} \) denotes the inverse variance of the reanalysis error for the \( j \)-th latent variable at time step \( i \), serving as a weighting factor in the loss.

However, as the true atmospheric state is unavailable, the reanalysis uncertainty in latent space is difficult to quantify directly. One possible approximation is to estimate it from the ERA5 analysis ensemble~\cite{hersbachERA5GlobalReanalysis2020}, but this would substantially increase computational cost and implementation complexity. Here, we cast the optimization across latent variables as a multitask learning problem and adopt a multitask negative log-likelihood (NLL) objective~\cite{cipollaMultitaskLearningUsing2018a} to learn the weighting parameters automatically. 
Specifically, we assume that the forecast error of each latent variable follows an independent zero-mean Gaussian distribution with variance $\sigma_{j,i}^2$. Under this assumption, we parameterize the corresponding log-variance, $\log \sigma_{j,i}^2$, and jointly optimize it with the model parameters using the following loss function:
\begin{equation}
\begin{aligned}
J(\boldsymbol{\theta})
&= \sum_{i=1}^{T} \sum_{j=1}^{m} \frac{\varepsilon_{j,i}^2}{\sigma_{j,i}^2}
+ \sum_{i=1}^{T} \sum_{j=1}^{m} \log \sigma_{j,i}^2, \\
\varepsilon_{j,i}
:&= \mathcal{E}^{(j)}(\boldsymbol{x}_{a,i})
- \mathcal{E}^{(j)}\!\left(\mathcal{M}_{0 \rightarrow i}(\boldsymbol{x}_{a,0}, \boldsymbol{\theta})\right).
\end{aligned}
\label{eq:nll_latent_loss}
\end{equation}
This NLL loss replaces hand-tuned weights with data-driven uncertainty estimates, and can likewise be used in place of the model-space MSE in Eq.~\ref{eq:weighted_mse}.

These derivations show that DFM training can enforce complex multivariate consistency in the high-dimensional model space through a simple latent-space MSE objective, provided that three conditions are satisfied: 1) the AE has a small reconstruction error; 2) the AE encoding–decoding process is approximately affine during training; and 3) the error covariance is approximately diagonal in the AE latent space.
Fig.~\ref{fig:fig1}A illustrates the distinction between DFM training under model-space and latent-space losses. For clarity, we denote the DFM trained with latent-space constraints as the DFM-LC, and the model trained with model-space constraints as the DFM-MC.

\section{Results}

We validate the proposed framework by comparing DFMs trained with different loss functions on the same coarsened ERA5 reanalysis~\cite{hersbachERA5GlobalReanalysis2020}, on a global 128 $\times$ 256 latitude-longitude grid (1.40625$^\circ$ resolution).
We use 69 variables to define the model state for forecasting, comprising five upper-air variables specified at 13 pressure levels (50, 100, 150, 200, 250, 300, 400, 500, 600, 700, 850, 925, and 1000 hPa) and four surface variables. The upper-air variables include geopotential height (Z), temperature (T), zonal wind (U), meridional wind (V), and specific humidity (Q). The surface variables are 2-meter air temperature (T2M), 10-meter zonal and meridional winds (U10 and V10), and mean sea level pressure (MSL).

We use the same AE and DFM architectures as in previous work on latent DA for the global atmosphere~\cite{fanPhysicallyConsistentGlobal2026}, with both models built using Swin Transformers~\cite{liuSwinTransformerHierarchical2021}. The DFM follows the design of Fengwu~\cite{chenOperationalMediumrangeDeterministic2025}, an operational medium-range forecasting model with a 6-h forecast step. Unlike most ML-based forecast models~\cite{lamLearningSkillfulMediumrange2023,chenFuXiCascadeMachine2023,kochkovNeuralGeneralCirculation2024,liGenerativeEmulationWeather2024,chenOperationalMediumrangeDeterministic2025},  which take atmospheric states from two consecutive time steps as input, the employed DFM uses only a single time step, aligning with the formulation of 4DVar. The AE compresses the full atmospheric state into a latent representation of dimensions 34×32×64, yielding a compression ratio of approximately 32.

For training DFM-LC, we use the AE parameters from~\cite{fanPhysicallyConsistentGlobal2026}, which satisfy the assumptions underlying the derivation of Eq.~\ref{eq:full_cov_eigen}. Specifically, the decoder was shown to be locally approximately affine (Fig. S1), while the background error covariance in the learned latent space was found to be nearly diagonal. Fig.~\ref{fig:fig1}B and C further compare forecast-error correlations in model space and latent space, providing additional evidence that latent-space errors are nearly decorrelated. Given the AE decoder is only approximately affine in the vicinity of the data manifold~\cite{fanPhysicallyConsistentGlobal2026}, we adopt a two-stage training strategy. 
In the first stage, the model is trained for 50 epochs without autoregressive rollout using a model-space NLL loss, which brings the model predictions closer to the data manifold and enables the subsequent application of latent-space constraints.
In the second stage, the model is fine-tuned using Eq.~\ref{eq:nll_latent_loss}, with the rollout length gradually increased from 2 to 12 steps following the curriculum learning strategy used in GraphCast~\cite{lamLearningSkillfulMediumrange2023}. Each rollout length is trained for one epoch, except for the final 12-step rollout, which is trained for two epochs to ensure convergence. The learning rate in the first stage is set to 5×$10^{-4}$ with warm-up and cosine decay scheduling, and fixed at 3×$10^{-7}$ in the second stage. DFM-MC is trained using the same two-stage strategy, but the NLL loss is computed solely in model space throughout. We note that training DFM-LC requires more computational resources than DFM-MC, with the AE encoder involved in the loss computation. Specifically, if the forecasting model has $A$ parameters and the AE encoder has $B$, the tuning cost of DFM-LC is approximately $(1 + B/A)$ times that of DFM-MC.

\begin{figure*}[ht]
    \centering
    \includegraphics[width=1.0\linewidth]{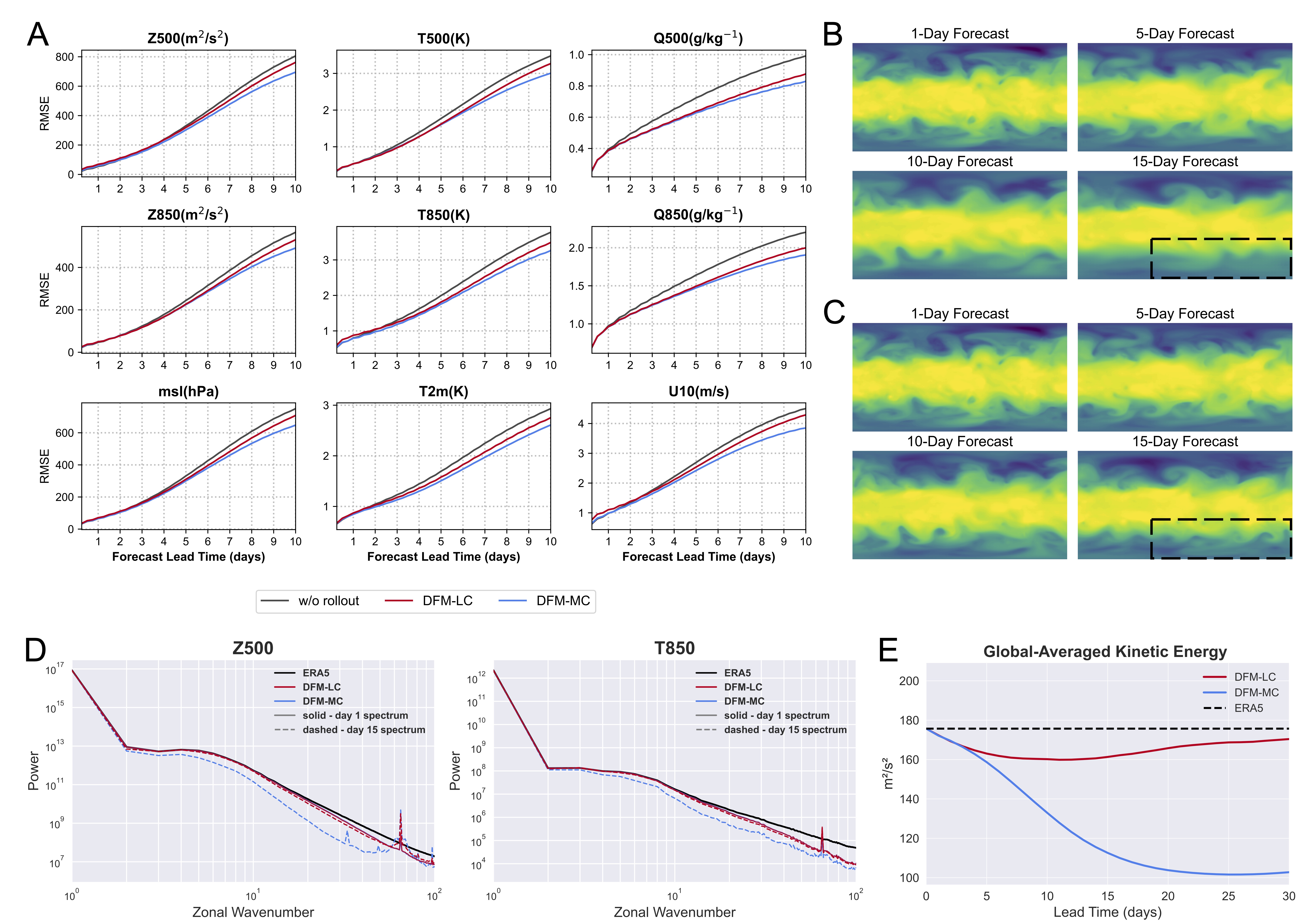}
    \caption{
    Forecast accuracy and spectral characteristics of DFM-LC and DFM-MC.
    (A) Averaged forecast errors of DFM-LC, DFM-MC, and the DFM trained without rollout. Forecasts are initialized twice daily from ERA5 reanalysis throughout 2020 and evaluated against ERA5 using latitude-weighted root mean square error.
    (B and C) Example T500 forecasts initialized from ERA5 reanalysis at 0000 UTC 1 Jan 2020, with black boxes highlighting differences in fine-scale structures at the 15-d lead.
    (D) Zonal-mean power spectra of Z500 and T850 over the midlatitudes (30°–60° N/S), computed from forecasts at day 1 (solid lines) and day 15 (dashed lines), and compared with ERA5.
    (E) Forecast evolution of free-atmospheric horizontal kinetic energy from DFM-LC and DFM-MC. Kinetic energy is computed as the mean horizontal wind energy across all grid points and pressure levels from 850 to 100 hPa. Results are averaged over forecasts initialized daily at 0000 UTC throughout 2020. The ERA5 reanalysis value is shown as a reference (dashed line).
    }
    \label{fig:fig2}
\end{figure*}

\subsection{Accuracy and Spectral Analysis}

We first assess the forecast accuracy of different DFMs initialized from and evaluated against the ERA5 reanalysis. Following the WeatherBench setup~\cite{raspWeatherBench2Benchmark2024}, the forecasts are initialized at 00 and 12 UTC each day throughout 2020, from 00 UTC on January 1 to 12 UTC on December 31, and are evaluated using the latitude-weighted root mean square error~\cite{biAccurateMediumrangeGlobal2023}. Fig.~\ref{fig:fig2}A compares forecast errors across several representative variables for models trained with different loss functions. Rollout training with either latent-space or model-space constraints substantially improves long-term forecast skill.
Over the first five lead days, DFM-LC remains close to DFM-MC, with relative RMSE differences below 2\% for 48 variables. By comparison, the forecast model trained without rollout shows relative RMSE differences below 2\% from DFM-MC for only six variables.
At longer lead times, however, DFM-MC consistently outperforms DFM-LC across all variables.

Although the DFM-MC achieves lower RMSE in long-range forecasts, its outputs appear noticeably more blurry than those of the DFM-LC, as illustrated by the case in Fig.~\ref{fig:fig2}B and C. To quantify this blurring effect, we calculate the zonal average power spectra of forecasts for geopotential height at 500 hPa (Z500) and temperature at 850 hPa (T850) over the midlatitudes (30°–60° N/S). 
As shown in Fig.~\ref{fig:fig2}D, the spectral distributions of all DFMs are consistent and closely approximate those of ERA5 at a 1-day lead time. However, with a 15-day lead time, the DFM-MC exhibits substantial loss of energy across a broad range of wavenumbers, indicating degraded spatial structure. 
In contrast, the DFM-LC retains significantly more spectral energy, even at 15 days, demonstrating its ability to preserve fine-scale features in extended forecasts.

A deficit of energy at high wavenumbers in the wind spectrum typically reflects excessive kinetic energy dissipation, potentially weakening circulations and eroding mesoscale structure.
We therefore quantify free-tropospheric kinetic energy dissipation by computing the mean horizontal kinetic energy across all grid points and pressure levels from 850 to 100 hPa. Fig.~\ref{fig:fig2}E shows the evolution of predicted kinetic energy from day 1 to day 30, averaged over all forecasts initialized daily at 00 UTC throughout 2020, for DFM-MC and DFM-LC. DFM-MC shows a rapid decline, losing 27.2\% of its initial kinetic energy by day 10 and more than 41.3\% by day 20. By contrast, DFM-LC stays much closer to ERA5, with a mean loss of only 3.6\% over the forecast period. This indicates that DFM-LC preserves horizontal kinetic energy much more effectively than DFM-MC.


\subsection{Cross-Scale Dynamics} 

In the real atmosphere, energy at large scales cascades toward smaller scales through nonlinear interactions. Therefore, a realistic atmospheric model should be able to regenerate fine-scale structures even when they are absent from the initial conditions. To assess this capability in DFMs, we design a forecast experiment in which all spectral components with total wavenumber greater than 6 are removed in the initial fields. We repeat this experiment daily throughout 2020 and focus on Z500, which provides a representative measure of free-atmospheric dynamics.

\begin{figure*}[ht]
    \centering
    \includegraphics[width=1.0\linewidth]{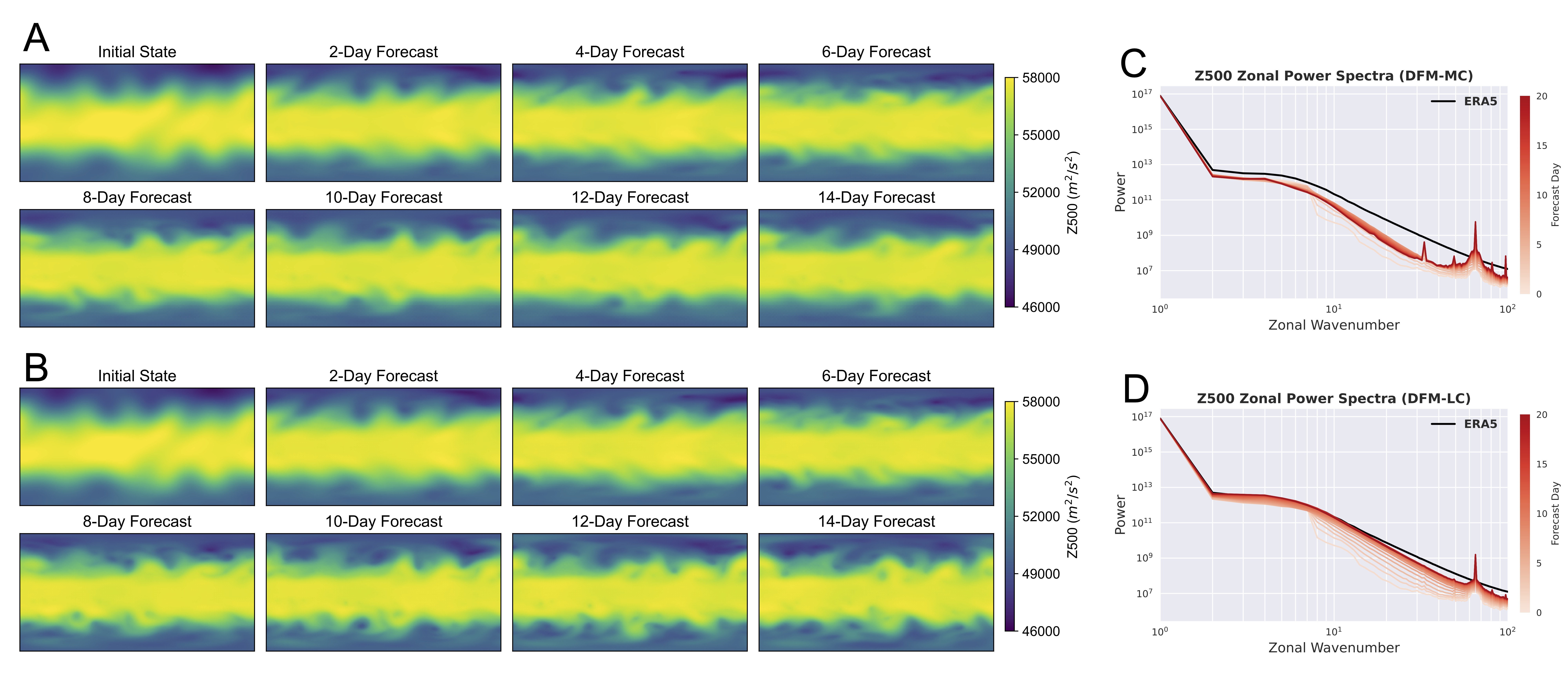}
    \caption{
    Forecasts from DFM-MC and DFM-LC initialized from ERA5 reanalysis with wavenumbers greater than 6 smoothed out.
    (A and B) Example Z500 forecasts from DFM-MC (A) and DFM-LC (B), initialized at 0000 UTC 1 Jan 2020.
    (C and D) Evolution of the zonal power spectra of Z500 forecasts from DFM-MC (C) and DFM-LC (D), averaged over experiments initialized daily at 0000 UTC throughout 2020.
    }
    \label{fig:fig3}
\end{figure*}

We find that DFM-LC and DFM-MC behave similarly at early lead times but differ markedly in their ability to regenerate fine-scale structure at longer lead times, as illustrated by a representative case in Fig.~\ref{fig:fig3}A and B. 
To quantify this behavior, we compute the evolution of the zonal Z500 spectra in the forecasts of these two DFMs, averaged over daily experiments in 2020 (Fig.~\ref{fig:fig3}C and D). Power at higher wavenumbers increases gradually with lead time in both forecasts, but the spectrum of DFM-LC converges toward a distribution that is closer to ERA5 than that of DFM-MC. Fig.~S2 in Supporting Information further reports the fraction of Z500 spectral power at wavenumbers larger than 6 relative to the total Z500 spectral power. For DFM-MC, this fine-scale fraction increases during the first 6 forecast days and then saturates at approximately $0.5\times 10^{-3}$, which is far below the ERA5 reference value of $2.1\times 10^{-3}$. In contrast, this fine-scale fraction increases throughout the first 14 forecast days in DFM-LC and then stabilizes close to the corresponding fraction in ERA5. These results indicate that DFM-LC learns more realistic dynamics than DFM-MC, supporting the generation of more realistic fine-scale features.

Interestingly, the forecasts of DFM-MC and DFM-LC remain highly similar over the first 6 forecast days of this experiment, which correspond to an adjustment stage from smoothed initial conditions that are physically unrealistic. During this period, the correlation between the two forecasts reaches 0.95, indicating their similar large-scale dynamical behavior. Thereafter, this consistency drops rapidly as DFM-LC begins to generate finer-scale structures that interact with the large-scale flow. This suggests that the forecast differences between DFM-LC and DFM-MC primarily stem from the way the learned dynamics represent interactions between large and small scales. Notably, the DFM-LC develop fine-scale structures primarily after the establishment of the large-scale circulation, rather than simultaneously, which is dynamically consistent with the multiscale character of atmospheric dynamics~\cite{hoskins1972atmospheric,hoskins1982mathematical}.

\subsection{Realism of Geostrophic Balance} 
\begin{figure*}[ht]
    \centering
    \includegraphics[width=1.0\linewidth]{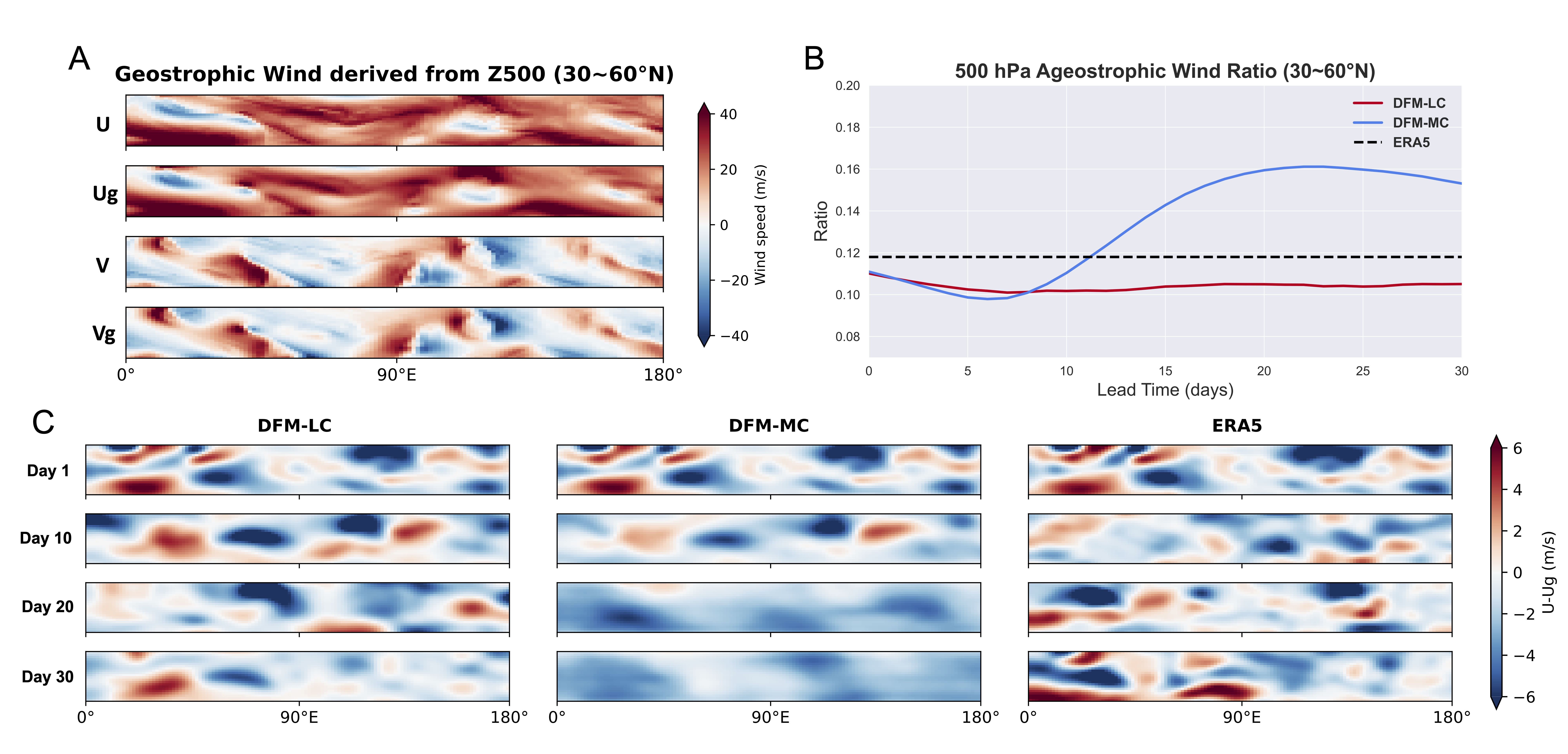}
    \caption{
    Geostrophic balance diagnostics for forecasts from DFM-LC and DFM-MC.
    (A) Zonal and meridional components of the actual wind $(u, v)$ and geostrophic wind $(u_g, v_g)$ at 500 hPa over the Northern Hemisphere midlatitudes (30$^\circ$--60$^\circ$N), derived from ERA5 reanalysis at 0000 UTC 1 January 2020.
    (B) Evolution of the geostrophic imbalance ratio $R_{\mathrm{imb}}$ over 30-day forecasts from DFM-LC and DFM-MC, averaged over forecasts initialized daily at 0000 UTC throughout 2020.
    (C) Zonal distribution of the difference between actual and geostrophic zonal wind at 500 hPa for forecasts initialized at 0000 UTC 1 January 2020, shown at selected lead times (days 1, 10, 20, and 30) for DFM-LC and DFM-MC, with ERA5 shown for reference.
    }
    \label{fig:fig4}
\end{figure*}

To examine whether latent-space constraints improve physical realism, we evaluate geostrophic balance in DFM forecasts. As one of the most fundamental balances in the midlatitude free atmosphere, geostrophic balance describes the equilibrium between the Coriolis force and the horizontal pressure-gradient force, yielding the diagnostic relation~\cite{IntroductionDynamicMeteorology2013}
\begin{equation}
\begin{aligned}
u_g = -\frac{1}{f} \frac{\partial \Phi}{\partial y},  v_g =\frac{1}{f} \frac{\partial \Phi}{\partial x},
\end{aligned}
\end{equation}
where $u_g$ and $v_g$ denote the zonal and meridional components of the geostrophic wind, respectively, $f$ is the Coriolis parameter, and $\Phi$ is the geopotential height. 
As shown in Fig.~\ref{fig:fig4}A, at 500~hPa over the Northern Hemisphere midlatitudes, the diagnosed geostrophic wind closely matches the actual ERA5 wind, indicating that geostrophic balance dominates the large-scale flow.

We assess the realism of geostrophic balance over this region in DFM forecasts using the relative magnitude of the ageostrophic component, which arises from flow curvature, diabatic forcing, and transient acceleration, defined as:
\begin{equation}
R_{\mathrm{imb}} = 
\frac{
\sqrt{(u - u_g)^2 + (v - v_g)^2}
}{
\sqrt{u^2 + v^2}
}.
\end{equation}
Fig.~\ref{fig:fig4}B shows the mean imbalance ratio over 1–30-day forecast lead times in this region, averaged over forecasts initialized at 00 UTC each day in 2020 for both the DFM-MC and DFM-LC. The DFM-LC maintains a stable imbalance ratio over time, with values slightly lower than those of ERA5 (by approximately 1.7\% on average). In contrast, although the DFM-MC forecasts also generally exhibit geostrophic balance over the target region, their imbalance ratio varies over time and is less stable.

Fig.~\ref{fig:fig4}C illustrates a case that explains why the imbalance component in the DFM-MC becomes unstable over time. For the DFM-LC, the $u_g$ forecasts consistently exhibit alternating easterly and westerly patterns that closely resemble those in ERA5. By comparison, the DFM-MC produces persistent easterly $u_g$ in forecasts beyond day 15, which is physically implausible. This contrast, observed in almost all cases, highlights the importance of enforcing multivariate consistency during DFM training to preserve physical consistency in forecasts.

\subsection{Evaluating DFMs through State-Only 4DVar}
If model-parameter errors are neglected (i.e., the forecast model is assumed to be perfect), Eq.~\ref{eq:4dvar_full} reduces to the commonly used state-only 4DVar formulation (hereafter, SO-4DVar), in which the forecast model provides a hard dynamical constraint within the assimilation window for optimization of the background state~\cite{courtierStrategyOperationalImplementation1994,evensenDataAssimilationFundamentals2022}. Within this framework, for fixed observations and assimilation settings, improved DA performance in estimating the model state implies that the forecast model more closely satisfies the perfect-model assumption.

Based on this principle, we assess the dynamical fidelity of different DFMs through the SO-4DVar analyses they support under idealized conditions. In this evaluation, ERA5 reanalysis is treated as truth, and synthetic observations are sampled every 12 h throughout 2017 at the locations of real surface and radiosonde observations.
For each DFM, we construct a cycling DA--forecast system (DAFS) that assimilates these observations using SO-4DVar and then evaluates both analyses and the forecasts initialized from each analysis against ERA5. To simplify estimation of the background error covariance matrix, we perform SO-4DVar in latent space, as in ~\cite{fanPhysicallyConsistentGlobal2026}. Further methodological details and experimental design are provided in the Supporting Information.

\begin{figure*}[ht]
    \centering
    \includegraphics[width=0.9\linewidth]{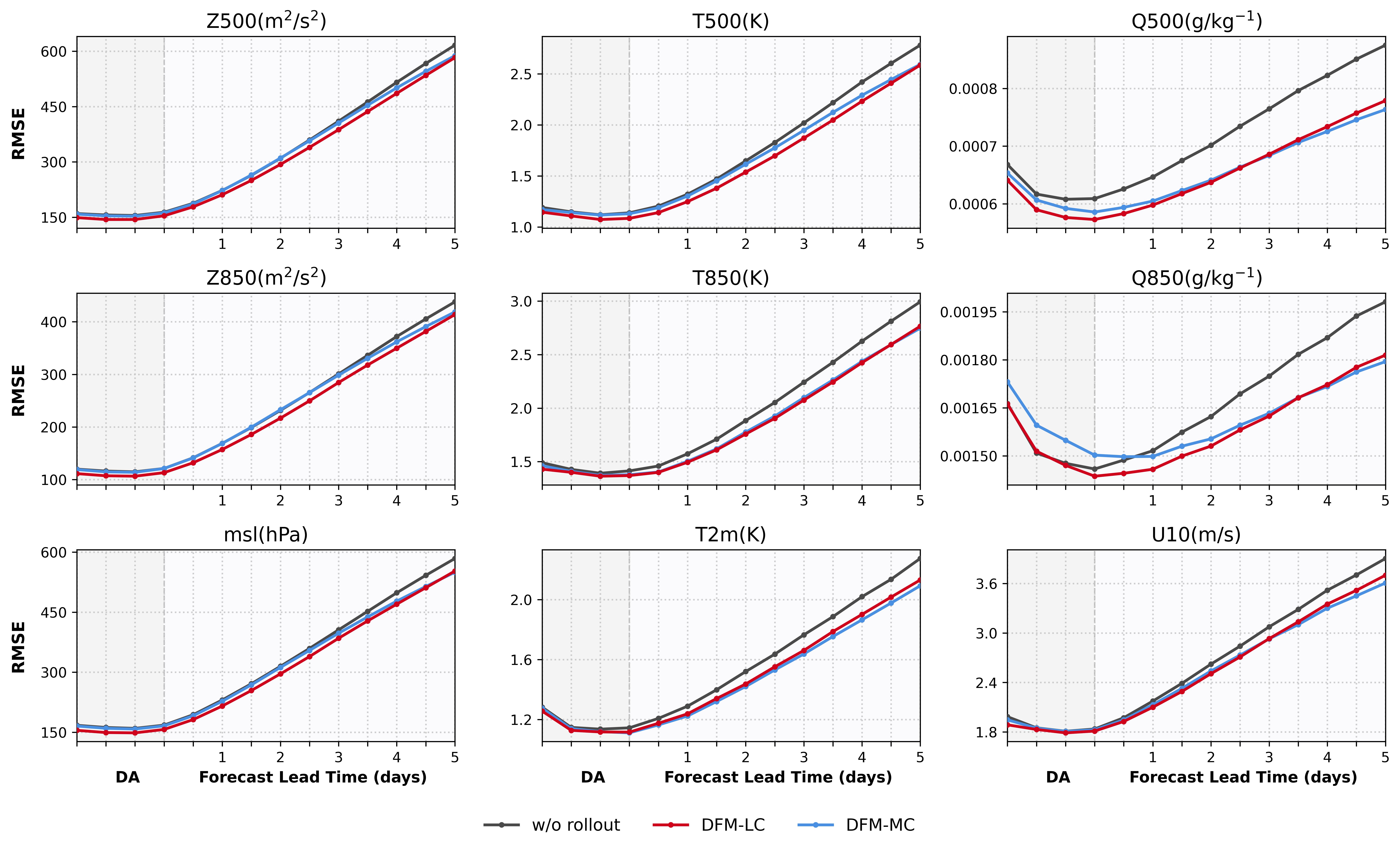}
    \caption{
    Analysis and forecast errors of cycling DA systems supported by different DFMs in idealized experiments, using the 4DVar formulation in which only the model state is optimized. Observations are sampled from ERA5 reanalysis at 12-h intervals throughout 2017 using the locations of real surface and radiosonde observations. Errors are evaluated against ERA5 using latitude-weighted root mean square error and then averaged over all DA cycles in 2017.
    }
    \label{fig:fig5}
\end{figure*}

Fig.~\ref{fig:fig5} shows the mean analysis and forecast errors for each DA cycle in the idealized experiments throughout 2017. Both the analysis and forecast skill of the DAFS improve substantially when the underlying DFM is trained with the rollout strategy, highlighting the importance of rollout for learning more faithful dynamics. Surprisingly, although the forecast skill of DFM-LC remains lower than that of DFM-MC, the DAFS supported by DFM-LC produces more accurate analyses than that supported by DFM-MC for 66 of 69 variables, with a 2.7\% reduction in analysis error when averaged over all variables. This advantage persists into the forecast stage, as the DAFS supported by DFM-LC outperforms that supported by DFM-MC for more than half of the variables over the first 4 forecast days. These results show that DFM-LC yields more accurate analyses than DFM-MC within the SO-4DVar framework, indicating that the dynamics learned by DFM-LC are more consistent with those represented in ERA5 reanalysis.

We note that this conclusion is difficult to obtain in real-observation experiments because observation errors cannot be estimated as precisely as in the idealized setting. When using model dynamics as a strong constraint, biased estimation of observation errors can lead to biased analyses even when the forecast model is perfect. 
Fig.~S3 shows that, when assimilating real observations, the analysis skill of DAFS supported by different DFMs differs only marginally in estimating the analysis fields, and the forecast errors of these systems closely track the forecast skill of the underlying DFMs.
Under this setting, it is therefore difficult to determine which DFM better captures the true atmospheric dynamics. Nevertheless, these experiments indicate that DFMs trained with different strategies can support SO-4DVar with real observations with comparable analysis accuracy.

\section{Toward a General Framework for DFM Training}
A key insight from DA is that, in principle, any source of information can improve state estimation, provided that its uncertainty is properly quantified~\cite{evensenDataAssimilationFundamentals2022}. This principle underlies modern operational DA systems, which integrate diverse observations to reconstruct past atmospheric states and support accurate weather prediction~\cite{carrassiDataAssimilationGeosciences2018}. Likewise, DFM training should benefit from exploiting as many complementary sources of information as possible to obtain the best estimate of atmospheric dynamics, rather than relying solely on reanalysis or observations.
We therefore propose a general framework for DFM training from a 4DVar perspective, in which multiple sources of information are jointly exploited. 

Assuming that multiple observation sources are available and that observation errors are uncorrelated in time and across observation types, Eq.~\ref{eq:theta_loss} can be reformulated as:
\begin{equation}
J(\boldsymbol{\theta})
= \sum_{i=1}^{T} \sum_{j=1}^{K}
\left\| \mathbf{y}_{i,j}
- \mathcal{H}_j\!\left(
\mathcal{M}_{0 \rightarrow i}(\boldsymbol{x}, \boldsymbol{\theta})
\right)
\right\|^2_{\mathbf{R}_{i,j}^{-1}} ,
\label{eq:4dvar_multi_source}
\end{equation}
where \( K \) denotes the number of independent observation sources, 
\( j = 1, \dots, K \) indexes the source type, 
\( \mathbf{y}_{i,j} \) is the observation from source \( j \) at time \( i \), 
\( \mathcal{H}_j \) is the corresponding observation operator, 
and \( \mathbf{R}_{i,j} \) is the associated observation error covariance matrix.

We group available observations for the atmosphere into two broad categories: discrete and continuous. For discrete observations, such as surface stations and radiosondes, the observation operator can often be constructed by simple interpolation, and the observation error covariance can be reasonably approximated as diagonal. The corresponding loss can therefore be formulated as a weighted MSE in observation space. By contrast, for continuous ``observations'', such as reanalysis fields, model forecasts, and satellite retrievals, the observation operator can be defined through interpolation or learned with neural networks. Their error covariance matrices, however, are typically high-dimensional, spatially structured, and difficult to estimate. 

An effective way to address this challenge is to compute the loss for each type of continuous observation in its latent space learned by AE, where errors are substantially less correlated than in the original space, as shown in this study. In this way, both discrete and continuous observations can be formulated as MSE losses in different spaces, and Eq.~\ref{eq:4dvar_multi_source} can be rewritten as follows:
\begin{equation}
J(\boldsymbol{\theta})
= \sum_{i=1}^{T} \sum_{d=1}^{D}
\left\| \boldsymbol{w}_{d,i} \odot
\left( \mathbf{y}_{d,i} - \widehat{\mathbf{y}}_{d,i} \right) \right\|^2
+ \sum_{i=1}^{T} \sum_{c=1}^{C}
\left\| \boldsymbol{k}_{c,i} \odot
\left( \mathcal{E}_{c}(\mathbf{y}_{c,i})
- \mathcal{E}_{c}(\widehat{\mathbf{y}}_{c,i}) \right) \right\|^2 ,
\label{eq:general_loss}
\end{equation}
where \( D \) and \( C \) are the numbers of discrete and continuous observation types, respectively; 
\( \widehat{\mathbf{y}}_{\cdot,i} = \mathcal{H}_{\cdot} ( \mathcal{M}^{(\cdot)}_{0 \rightarrow i}(\mathbf{x}_{a,0}, \boldsymbol{\theta}) ) \) 
is the model forecast mapped to the corresponding observation space; 
\( \mathcal{E}_c \) is the encoder for the \( c \)-th continuous observation type; and \( \odot \) denotes element-wise multiplication.
\( \boldsymbol{w}_{d,i} \) and \( \boldsymbol{k}_{c,i} \) are variable-wise weighting vectors for the discrete and continuous terms, respectively, and can be either prescribed manually or learned jointly with the model through the NLL loss.

We emphasize that the assumptions used to derive Eq.~\ref{eq:weighted_latent_loss} should be verified before replacing the loss for continuous observations with an MSE loss in latent space. For observation types that do not satisfy these conditions, they can instead be treated as discrete observations with mutually independent errors. Moreover, as in traditional DA, incorporating multiple observational sources often requires empirical tuning to achieve optimal performance.

\section{Discussion}
In this study, we reformulate the training of deterministic ML-based weather forecasting models as a 4DVar problem. This perspective reveals that the widely used MSE loss ignores the covariance component of error statistics, potentially weakening the enforcement of physical consistency during training.
To address this issue, we instead compute the MSE loss in latent space, where the learned representation implicitly captures the complex multivariate consistency in atmosphere and thereby removes the need for explicit error-covariance estimation to enforce physical consistency.
Our results indicate that applying latent-space constraints during rollout training enhances long-range forecast accuracy, yields outputs that are more physically realistic, and better preserves fine-scale structures than model-space constraints.
Finally, we extend this framework to a more general formulation that allows DFMs to be trained with diverse observational sources under a unified objective.

From a frequentist perspective, the model-space MSE loss corresponds to a maximum-likelihood estimation under independent, zero-mean Gaussian forecast–reanalysis errors~\cite{bishop2006pattern}. We here adopted a Bayesian DA perspective instead for interpreting the training of an ML-weather forecast model because it provides a well-defined framework for estimating the quantities that characterize a dynamical system and its constraints under uncertainty ~\cite{evensenDataAssimilationFundamentals2022}. This viewpoint yielded two key insights. 
First, the derivation showed that DFM training implicitly assumes perfect initial states. This may be reasonable for parts of the atmosphere, where high-quality reanalyses are available, but may be problematic in domains or regions where observations are too sparse to support reliable state estimation.
Second, our derivation is essentially a so-called strong-constraint 4DVar formulation~\cite{evensenDataAssimilationFundamentals2022}, which neglects stochastic model error arising from missing physics and numerical discretization errors. Such intrinsic model uncertainty can accumulate in deterministic forecasts and eventually lead to large long-range errors, even when the learned parameters capture physically realistic dynamics. Therefore, extending this Bayesian framework to account for stochastic model error and ensemble prediction will be important for advancing long-range and probabilistic forecasting.

Approximating a covariance-aware model-space loss with a latent-space MSE objective is effective and practically simple, but it requires three conditions: small AE reconstruction error, local linearity of the decoder along the optimization trajectory, and an approximately diagonal error covariance in latent space. These conditions are difficult to guarantee simultaneously through AE architecture design and are unlikely to hold exactly in practice, which may limit the general applicability of the approach and inevitably introduce some additional bias into DFM training.
However, existing studies suggest that these conditions are often approximately satisfied across spatial scales, variables and even oceanic settings~\cite{fanNovelLatentSpace2025a,zhengGeneratingUnseenNonlinear,fanPhysicallyConsistentGlobal2026,melinc3DVarDataAssimilation2024,melincUnifiedNeuralBackgroundError2026}, possibly because geophysical data manifolds are smoother and more continuous than those in computer vision, thereby supporting the strong performance of latent DA.
We therefore believe that the latent-space constraints may have broad applicability in geoscientific research and other chaotic spatiotemporal systems.

We note that learning atmospheric dynamics with latent-space constraints is conceptually aligned with the emerging idea of world models~\cite{haWorldModels2018, assranVJEPA2SelfSupervised2025, hafnerMasteringDiverseControl2025}, which emphasize predicting the evolution of a complex system in the representation space rather than matching the raw data element by element. As argued by Yann LeCun~\cite{lecun2022path}, this strategy allows models to capture the structured, predictable components of a system while filtering out low-level details that are irrelevant or inherently unpredictable, which in atmospheric modeling correspond to stochastic model and observational error.
This perspective clarifies the principle by which latent-space constraints promote physically realistic dynamics, and suggests that future Earth system prediction may benefit from advances in world models.

\section{Acknowledgments}
H. F., Y. Q., J. N., and P. G. acknowledge support from the National Science Foundation (NSF) Science and Technology Center (STC) Learning the Earth with Artificial Intelligence and Physics (LEAP, Award \#2019625).

\bibliographystyle{unsrt}

\bibliography{references}  






\clearpage
\newpage

\section*{Supplementary materials}
\setcounter{figure}{0}
\setcounter{equation}{0}
\renewcommand{\thefigure}{S\arabic{figure}}

\section*{Derivation of four-dimensional variational data assimilation}
We consider a standard strong-constraint four-dimensional variational data assimilation (4DVar) problem over an assimilation window
$t_0,t_1,\dots,t_T$, in which the initial state $\boldsymbol{x}$ and time-invariant model parameters $\boldsymbol{\theta}$ are optimized simultaneously.
We assume that, within the assimilation window, the forecast model is deterministic and is fully driven by $\boldsymbol{x}$ and $\boldsymbol{\theta}$.
Accordingly, the model state at time $t_i$ is given by
\begin{equation}
\boldsymbol{x}_i
=
\mathcal{M}_{0\to i}(\boldsymbol{x},\boldsymbol{\theta}),
\label{eq:si_model_evolution}
\end{equation}
where $\mathcal{M}_{0\to i}$ denotes the nonlinear model propagator from $t_0$ to $t_i$.

At time $t_i$, the observation vector $\boldsymbol{y}_i$ is related to the model state through the observation operator $\mathcal{H}_i$:
\begin{equation}
\boldsymbol{y}_i
=
\mathcal{H}_i\!\left(\mathcal{M}_{0\to i}(\boldsymbol{x},\boldsymbol{\theta})\right)
+
\boldsymbol{\varepsilon}_i,
\label{eq:si_obs_equation}
\end{equation}
where the observation error is assumed Gaussian,
\begin{equation}
\boldsymbol{\varepsilon}_i \sim \mathcal{N}(\boldsymbol{0},\mathbf{R}_i).
\end{equation}
We further assume that observation errors are mutually independent in time and independent of the background errors.

The prior distributions of the initial state and model parameters are assumed to be Gaussian:
\begin{equation}
\boldsymbol{x}\sim\mathcal{N}(\boldsymbol{x}_b,\mathbf{B}),
\qquad
\boldsymbol{\theta}\sim\mathcal{N}(\boldsymbol{\theta}_b,\boldsymbol{\Theta}),
\label{eq:si_prior}
\end{equation}
where $\boldsymbol{x}_b$ and $\boldsymbol{\theta}_b$ are the background estimates, and
$\mathbf{B}$ and $\boldsymbol{\Theta}$ are the corresponding background error covariance matrices.
Assuming that the background errors of $\boldsymbol{x}$ and $\boldsymbol{\theta}$ are independent, the joint prior density can be written as
\begin{equation}
p(\boldsymbol{x},\boldsymbol{\theta})
\propto
\exp\left(
-\frac{1}{2}\|\boldsymbol{x}-\boldsymbol{x}_b\|_{\mathbf{B}^{-1}}^2
-\frac{1}{2}\|\boldsymbol{\theta}-\boldsymbol{\theta}_b\|_{\boldsymbol{\Theta}^{-1}}^2
\right),
\label{eq:si_joint_prior}
\end{equation}
where $\|\boldsymbol{v}\|_{\mathbf{W}}^2 \equiv
\boldsymbol{v}^{\mathrm T}\mathbf{W}\boldsymbol{v}$.

Given $\boldsymbol{x}$ and $\boldsymbol{\theta}$, the likelihood of the observation sequence
$\boldsymbol{y}_{1:T}=\{\boldsymbol{y}_1,\dots,\boldsymbol{y}_T\}$ is
\begin{equation}
p(\boldsymbol{y}_{1:T}\mid \boldsymbol{x},\boldsymbol{\theta})
=
\prod_{i=1}^{T}
p(\boldsymbol{y}_i\mid \boldsymbol{x},\boldsymbol{\theta}).
\end{equation}
Using Eq.~\ref{eq:si_obs_equation} and the Gaussian error assumption, this becomes
\begin{equation}
p(\boldsymbol{y}_{1:T}\mid \boldsymbol{x},\boldsymbol{\theta})
\propto
\exp\left(
-\frac{1}{2}\sum_{i=1}^{T}
\left\|
\boldsymbol{y}_i
-
\mathcal{H}_i\!\left(\mathcal{M}_{0\to i}(\boldsymbol{x},\boldsymbol{\theta})\right)
\right\|_{\mathbf{R}_i^{-1}}^2
\right).
\label{eq:si_likelihood}
\end{equation}

By Bayes' theorem, the posterior density is
\begin{equation}
p(\boldsymbol{x},\boldsymbol{\theta}\mid \boldsymbol{y}_{1:T})
\propto
p(\boldsymbol{y}_{1:T}\mid \boldsymbol{x},\boldsymbol{\theta})\,
p(\boldsymbol{x},\boldsymbol{\theta}).
\label{eq:si_bayes}
\end{equation}
Taking the negative logarithm of Eqs.~\ref{eq:si_joint_prior},~\ref{eq:si_bayes} and discarding terms independent of
$\boldsymbol{x}$ and $\boldsymbol{\theta}$ yields the 4DVar objective function
\begin{equation}
J(\boldsymbol{x},\boldsymbol{\theta})
=
\frac{1}{2}\|\boldsymbol{x}-\boldsymbol{x}_b\|_{\mathbf{B}^{-1}}^2
+
\frac{1}{2}\|\boldsymbol{\theta}-\boldsymbol{\theta}_b\|_{\boldsymbol{\Theta}^{-1}}^2
+
\frac{1}{2}\sum_{i=1}^{T}
\left\|
\boldsymbol{y}_i
-
\mathcal{H}_i\!\left(\mathcal{M}_{0\to i}(\boldsymbol{x},\boldsymbol{\theta})\right)
\right\|_{\mathbf{R}_i^{-1}}^2.
\label{eq:cost_4dvar_state_param}
\end{equation}

Minimizing Eq.~\ref{eq:cost_4dvar_state_param} therefore yields the maximum a posteriori (MAP) estimate of the initial state and model parameters.

\section*{Approximate Reformulation of the Model-Space Training Loss in Latent Space}

From the derivation in the main text, the training loss for a deterministic forecast model using reanalysis data is
\begin{equation}
J(\boldsymbol{\theta}) = \sum_{i=1}^{T} \left\| \boldsymbol{x}_{a,i} - \mathcal{M}_{0 \rightarrow i}(\boldsymbol{x}_{a,0}, \boldsymbol{\theta}) \right\|^2_{\mathbf{A}_i^{-1}},
\label{eq:theta_loss}
\end{equation}
where \( \boldsymbol{x}_{a,i} \) denotes the reanalysis state at time \(i\), and \( \mathbf{A}_i \) denotes the corresponding reanalysis error covariance matrix. However, \( \mathbf{A}_i \) is high-dimensional and difficult to estimate in practice.
To address this issue, we instead compute the training loss in a latent space learned by an autoencoder (AE), where the corresponding error covariance matrix admits a diagonal approximation.

In this section, we show that Eq.~\ref{eq:theta_loss} can be approximately reformulated in this latent space when the autoencoder satisfies the following conditions during model training:
\begin{itemize}
  \item[(1)] The AE reconstruction error is negligible on the data manifold, i.e., \( \mathcal{D}(\mathcal{E}(\boldsymbol{x})) \approx \boldsymbol{x} \), where \( \mathcal{E} \colon \mathbb{R}^n \to \mathbb{R}^m \) and \( \mathcal{D} \colon \mathbb{R}^m \to \mathbb{R}^n \) denote the encoder and decoder, respectively. Here, \( n \) is the dimensionality of the model (or physical) space, and \( m < n \) is the dimensionality of the latent space.
  \item[(2)] The encoder $\mathcal{E}$ and decoder $\mathcal{D}$ are both locally approximately affine in a neighborhood of the data manifold.
\end{itemize}

Under these assumptions, local deviations in model space and latent space are related through the encoder Jacobian \( \mathbf{J}_{\mathcal{E}} \) and decoder Jacobian \( \mathbf{J}_{\mathcal{D}} \) as
\begin{equation}
\boldsymbol{z}_1 - \boldsymbol{z}_2 \approx \mathbf{J}_{\mathcal{E}} (\boldsymbol{x}_1 - \boldsymbol{x}_2),
\quad
\boldsymbol{x}_1 - \boldsymbol{x}_2 \approx \mathbf{J}_{\mathcal{D}} (\boldsymbol{z}_1 - \boldsymbol{z}_2),
\label{eq:jacobian}
\end{equation}
where \( \boldsymbol{x}_1 \) and \( \boldsymbol{x}_2 \) are arbitrary nearby points on the data manifold, and \( \boldsymbol{z}_1=\mathcal{E}(\boldsymbol{x}_1) \) and \( \boldsymbol{z}_2=\mathcal{E}(\boldsymbol{x}_2) \). Substituting the first expression into the second yields, for an arbitrary local latent deviation \( \Delta \boldsymbol{z} = \boldsymbol{z}_1 - \boldsymbol{z}_2 \),
\begin{equation}
    \Delta \boldsymbol{z} \approx \mathbf{J}_{\mathcal{E}} (\mathbf{J}_{\mathcal{D}} \Delta \boldsymbol{z}) = (\mathbf{J}_{\mathcal{E}} \mathbf{J}_{\mathcal{D}}) \Delta \boldsymbol{z},
\end{equation}
suggesting that the Jacobians satisfy \( \mathbf{J}_{\mathcal{E}} \mathbf{J}_{\mathcal{D}} \approx \mathbf{I}_m \).

Eq.~\ref{eq:jacobian} also implies an approximate relationship between the model-space forecast error covariance \( \mathbf{A} \) and the latent-space error covariance \( \mathbf{A}_z \):
\begin{equation}
\begin{aligned}
\mathbf{A}
&= \mathbb{E} \left[ (\boldsymbol{x}_t - \boldsymbol{x}_a)(\boldsymbol{x}_t - \boldsymbol{x}_a)^\top \right] \\
&\approx \mathbb{E} \left[ \mathbf{J}_{\mathcal{D}} (\mathcal{E}(\boldsymbol{x}_t) - \mathcal{E}(\boldsymbol{x}_a))
(\mathcal{E}(\boldsymbol{x}_t) - \mathcal{E}(\boldsymbol{x}_a))^\top \mathbf{J}_{\mathcal{D}}^\top \right] \\
&= \mathbf{J}_{\mathcal{D}} \, \mathbb{E} \left[ (\mathcal{E}(\boldsymbol{x}_t) - \mathcal{E}(\boldsymbol{x}_a))
(\mathcal{E}(\boldsymbol{x}_t) - \mathcal{E}(\boldsymbol{x}_a))^\top \right] \mathbf{J}_{\mathcal{D}}^\top \\
&= \mathbf{J}_{\mathcal{D}} \, \mathbf{A}_z \, \mathbf{J}_{\mathcal{D}}^\top,
\label{eq:A_Az}
\end{aligned}
\end{equation}
where subscripts \( t \) and \( a \) denote the true state and the reanalysis (analysis) state, respectively. We note that this approximation is only valid locally on the decoded latent manifold, as the ranks of the left- and right-hand sides are determined by the dimensions of the model space and latent space, respectively.

By right-multiplying Eq.~\ref{eq:A_Az} by \( \mathbf{J}_{\mathcal{E}}^\top \), we obtain
\begin{equation}
\mathbf{A} \mathbf{J}_{\mathcal{E}}^\top
\approx
\mathbf{J}_{\mathcal{D}} \mathbf{A}_z (\mathbf{J}_{\mathcal{E}} \mathbf{J}_{\mathcal{D}})^\top
\approx
\mathbf{J}_{\mathcal{D}} \mathbf{A}_z,
\end{equation}
where we used \( \mathbf{J}_{\mathcal{E}} \mathbf{J}_{\mathcal{D}} \approx \mathbf{I}_m \). Left-multiplying both sides by \( \mathbf{J}_{\mathcal{D}}^\top \mathbf{A}^{-1} \) then gives
\begin{equation}
\mathbf{J}_{\mathcal{D}}^\top \mathbf{J}_{\mathcal{E}}^\top
\approx
\left( \mathbf{J}_{\mathcal{D}}^\top \mathbf{A}^{-1} \mathbf{J}_{\mathcal{D}} \right)\mathbf{A}_z.
\end{equation}
Using \( \mathbf{J}_{\mathcal{D}}^\top \mathbf{J}_{\mathcal{E}}^\top = (\mathbf{J}_{\mathcal{E}} \mathbf{J}_{\mathcal{D}})^\top \approx \mathbf{I}_m \), we obtain
\begin{equation}
\mathbf{I}_m
\approx
\left( \mathbf{J}_{\mathcal{D}}^\top \mathbf{A}^{-1} \mathbf{J}_{\mathcal{D}} \right)\mathbf{A}_z,
\end{equation}
and hence
\begin{equation}
\mathbf{J}_{\mathcal{D}}^\top \mathbf{A}^{-1} \mathbf{J}_{\mathcal{D}}
\approx
\mathbf{A}_z^{-1}.
\label{eq:inverse_relation}
\end{equation}

Now, for any analysis state \( \boldsymbol{x}_a \) and forecast state \( \boldsymbol{x} \) constrained near the manifold, the model-space quadratic form can be approximated by its latent-space counterpart:
\begin{equation}
\begin{aligned}
    (\boldsymbol{x} - \boldsymbol{x}_a)^\top \mathbf{A}^{-1} (\boldsymbol{x} - \boldsymbol{x}_a)
    &\approx (\mathbf{J}_{\mathcal{D}} (\boldsymbol{z} - \boldsymbol{z}_a))^\top \mathbf{A}^{-1} (\mathbf{J}_{\mathcal{D}} (\boldsymbol{z} - \boldsymbol{z}_a)) \\
    &= (\boldsymbol{z} - \boldsymbol{z}_a)^\top (\mathbf{J}_{\mathcal{D}}^\top \mathbf{A}^{-1} \mathbf{J}_{\mathcal{D}}) (\boldsymbol{z} - \boldsymbol{z}_a) \\
    &\approx (\mathcal{E}(\boldsymbol{x}) - \mathcal{E}(\boldsymbol{x}_a))^\top \mathbf{A}_z^{-1} (\mathcal{E}(\boldsymbol{x}) - \mathcal{E}(\boldsymbol{x}_a)),
\end{aligned}
\end{equation}
and thus the model-space loss in Eq.~\ref{eq:theta_loss} admits an approximate representation in latent space:
\begin{equation}
J(\boldsymbol{\theta}) = \sum_{i=1}^{T} \left\| \boldsymbol{x}_{a,i} - \mathcal{M}_{0 \rightarrow i}(\boldsymbol{x}_{a,0}, \boldsymbol{\theta})
\right\|^2_{\mathbf{A}_i^{-1}}
\approx \sum_{i=1}^{T} \left\| \mathcal{E}(\boldsymbol{x}_{a,i}) - \mathcal{E}(\mathcal{M}_{0 \rightarrow i}(\boldsymbol{x}_{a,0}, \boldsymbol{\theta}))
\right\|^2_{\mathbf{A}_{z,i}^{-1}}.
\label{eq:latent_theta_loss}
\end{equation}
We again emphasize that this approximation holds only on the manifold represented in latent space, rather than on the entire data manifold. Moreover, since the AE decoder that lifts latent states to model space is typically locally injective, the local affine properties of the decoder and encoder generally coexist on the decoded latent manifold. In practice, it is therefore often sufficient to verify only the local linearity of the decoder.

\section*{Setup of the data assimilation experiments}
\subsection*{Observations}
We generally follow the experimental design of Ref.~\cite{fanPhysicallyConsistentGlobal2026}. Global observations for the assimilation experiments are taken from GDAS for the year 2017. For simplicity, we assimilate only surface and radiosonde observations, corresponding to the BUFR codes ``ADPSFC'' and ``ADPUPA'', respectively. All observations are interpolated onto the model grid, and multiple observations within the same grid cell are averaged to reduce sampling noise. Surface observations located at elevations comparable to radiosonde levels are further assimilated as upper-air observations after adjustment to the nearest model pressure level using the hypsometric equation~\cite{IntroductionDynamicMeteorology2013}.

We performed cycling assimilation experiments throughout 2017 using both idealized and real observations. For the idealized-observation experiments, observations were generated by sampling ERA5 at the locations of the real observations and adding independent zero-mean Gaussian perturbations to each variable, with variance set to one-thousandth of the corresponding climatological variance. In the real-observation experiments, observations whose departures from ERA5 analyses exceed variable-specific thresholds are discarded to enhance assimilation stability. These thresholds are defined for each of the 69 variables as the mean absolute difference between ERA5 analyses separated by 48 hours, computed over the year 2016. To account for representativeness error and interpolation uncertainty associated with complex terrain, thresholds for surface observations are further increased by a factor of four. After preprocessing, the resulting dataset contains more than 3,000 surface observations and approximately 400 radiosonde observations every 12 hours.

\subsection*{DA method}
We adopt a latent-space 4DVar framework that is used only for state estimation, with the cost function given by:
\begin{equation}
J(\boldsymbol{z})
=
\frac{1}{2}
\left\|
\boldsymbol{z}-\boldsymbol{z}_b
\right\|_{\mathbf{B}_z^{-1}}^2
+
\frac{1}{2}
\sum_{i=0}^{T}
\left\|
\boldsymbol{y}_i
-
\mathcal{H}_i\!\left(
\mathcal{D}\!\left(
\mathcal{M}_{z,0\rightarrow i}(\boldsymbol{z}_0)
\right)
\right)
\right\|_{\mathbf{R}_i^{-1}}^2,
\label{eq:latent_4DVar}
\end{equation}
where $\boldsymbol{z}$ is the latent initial state to be estimated, and $\boldsymbol{z}_b$ is the latent representation of the background state $\boldsymbol{x}_b$. All other symbols have the same meanings as in Eq.~\ref{eq:cost_4dvar_state_param}. In our implementation, we use the same AE as that used for DFM training in the main text and in Ref.~\cite{fanPhysicallyConsistentGlobal2026}. Under this setting, the decoder is approximately locally linear and the latent background error covariance $\mathbf{B}_z$ is approximately diagonal, which makes the latent-space 4DVar formulation efficient to optimize. Ref.~\cite{fanPhysicallyConsistentGlobal2026} further showed that decoding the final latent analysis $\boldsymbol{z}_a$ yields an estimate that approximates the solution of the corresponding 4DVar problem in model space. In practice, Eq.~\ref{eq:latent_4DVar} is minimized with the Adam optimizer~\cite{kingmaAdamMethodStochastic2017}, since the inclusion of the nonlinear decoder and forecast model makes the objective function nonlinear.

\subsection*{Estimation of B matrix}
The latent-space background error covariance matrix $\mathbf{B}_z$ is estimated following the classical NMC method. In latent space, this estimation is substantially simplified because $\mathbf{B}_z$ is approximately diagonal. Specifically, each diagonal element is computed as
\begin{equation}
\mathbf{B}_{z,i} \approx \frac{1}{2} \left\langle \left( \mathcal{E}(\boldsymbol{x}^{48})_i - \mathcal{E}(\boldsymbol{x}^{24})_i \right)^2 \right\rangle,
\end{equation}
where $\boldsymbol{x}^{24}$ and $\boldsymbol{x}^{48}$ denote the 24-h and 48-h forecasts valid at the same time, respectively, $i$ indexes the latent dimensions, and $\langle \cdot \rangle$ denotes the average over samples.

To estimate $\mathbf{B}_z$, we computed $\boldsymbol{x}^{24}$ and $\boldsymbol{x}^{48}$ at 6-hourly intervals throughout 2016, yielding 1,460 forecast pairs. The resulting $\mathbf{B}_z$ was then used in the DA experiments for 2017. The magnitude of $\mathbf{B}_z$ was further tuned using experiments conducted in 2016. We note that $\mathbf{B}_z$ must be estimated separately for each forecast model.

\subsection*{Metrics}
For the idealized-observation experiments, we evaluate each atmospheric variable $c$ against ERA5 using the latitude-weighted root mean square error (RMSE):
\begin{equation}
\operatorname{RMSE}(\boldsymbol{x},\boldsymbol{x}_{\mathrm{truth}},c)
=
\sqrt{
\frac{1}{H W}
\sum_{h,w}
\frac{\cos\!\left(\alpha_{h,w}\right)}
{\sum_{h'=1}^{H}\cos\!\left(\alpha_{h',w}\right)}
\left(
\boldsymbol{x}^{c,h,w}
-
\boldsymbol{x}_{\mathrm{truth}}^{c,h,w}
\right)^2
},
\end{equation}
where the superscripts $c$, $h$, and $w$ denote the variable, latitude index, and longitude index, respectively. $\alpha_{h,w}$ is the latitude at grid point $(h,w)$. $H$ and $W$ are the numbers of grid points in the latitudinal and longitudinal directions, respectively.

For the real-observation experiments, validation observations are spatially sparse. We therefore measure the accuracy of the model state $\boldsymbol{x}$ using the root mean square error (RMSE) computed at the observation locations:
\begin{equation}
\operatorname{RMSE}(\boldsymbol{x},\boldsymbol{y},c)
=
\sqrt{
\frac{1}{N}
\sum_i
\left(
\mathcal{H}(\boldsymbol{x})^{c,i}
-
\boldsymbol{y}^{c,i}
\right)^2
},
\end{equation}
where the superscript $c,i$ denotes the $i$th observation of variable $c$, and $N$ is the total number of observations for that variable.

\begin{figure}[b]
    \centering
    \includegraphics[width=1.0\linewidth]{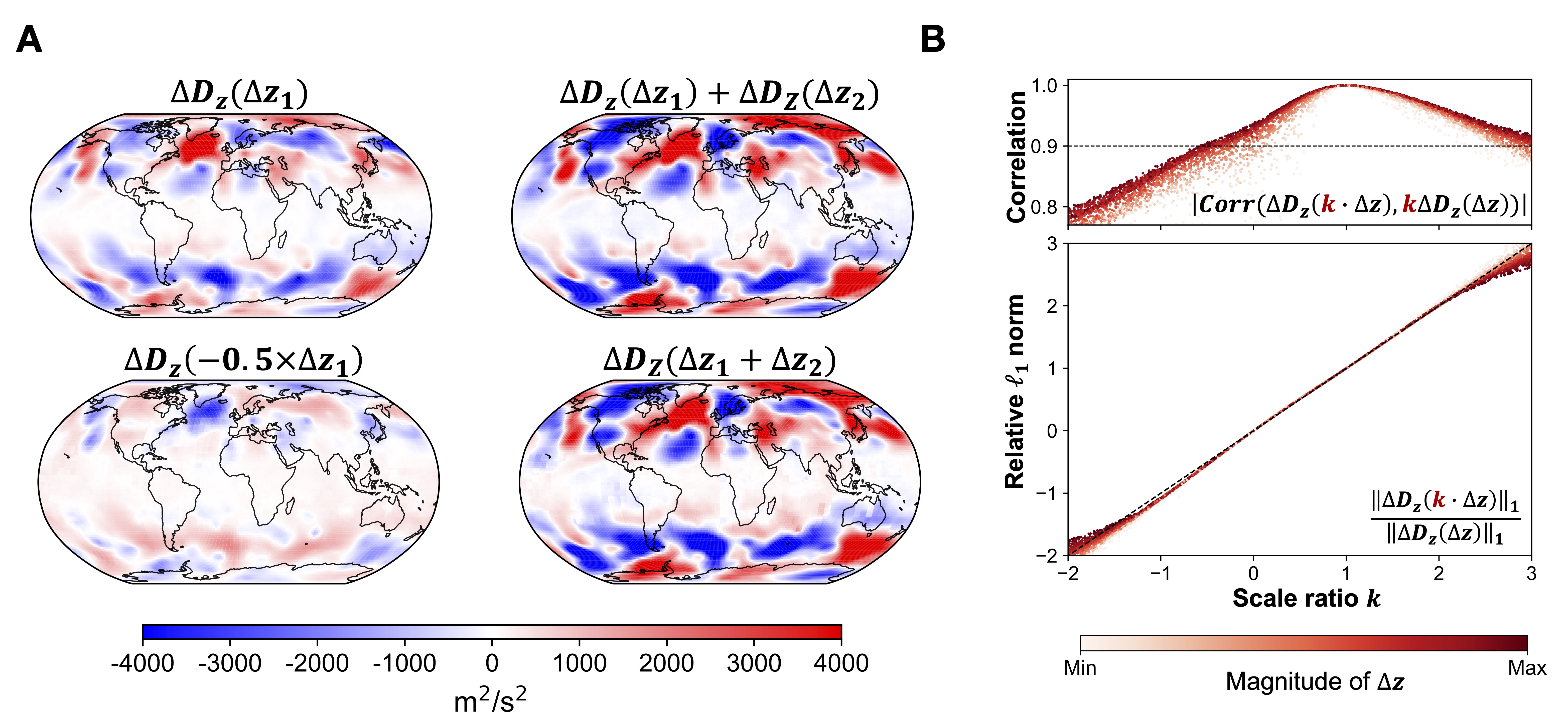}
    \caption{
    Approximately affine behavior of the AE decoder along latent directions representative of atmospheric variability. 
    (A), The impact of latent-space perturbations $\Delta \boldsymbol{z}$ at $\boldsymbol{z}$ on decoding results, denoted as $\Delta D_{\boldsymbol{z}}(\Delta \boldsymbol{z})$, shown using Z500. Here, $\boldsymbol{z}$ denotes the latent state corresponding to the ERA5 reanalysis at 0000 UTC on February 1, 2017. The perturbations $\Delta \boldsymbol{z_1}$ and $\Delta \boldsymbol{z_2}$ represent the latent differences between $\boldsymbol{z}$ and the reanalysis at 0000 UTC on January 1 and March 1, 2017, respectively. 
    (B), Evaluation of the near-linear response region of the AE decoder. The upper panel shows the correlation between $\Delta D_{\boldsymbol{z}}(k \cdot \Delta \boldsymbol{z})$ and $k \cdot \Delta D_{\boldsymbol{z}}(\Delta \boldsymbol{z})$ as a function of scale ratio $k$; the lower panel shows their relative $\ell_1$ norm.
    This figure is reproduced from~\cite{fanPhysicallyConsistentGlobal2026}, from which we adopt the same AE used in this study.
    }
\end{figure}

\begin{figure}[b]
    \centering
    \includegraphics[width=1.0\linewidth]{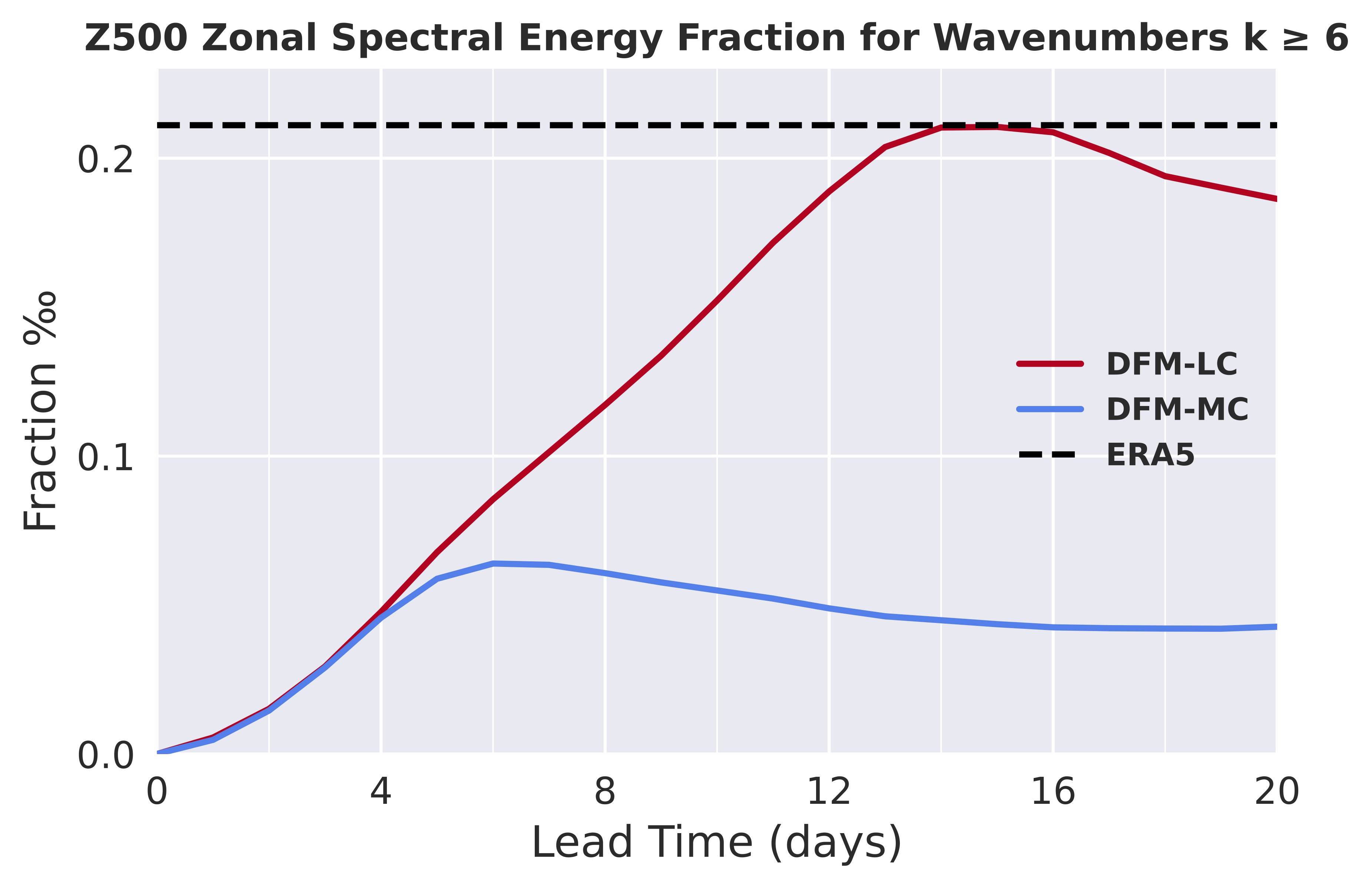}
    \caption{
    Recovery of high-wavenumber 500-hPa geopotential height in DFM forecasts for the experiments in Fig. 3 of the main text.
    Shown is the fraction of zonal spectral energy in Z500 associated with wavenumbers $k \geq 6$ , which are smoothed out in the initial states, for DFM-LC and DFM-MC as a function of lead time.
    }
\end{figure}

\begin{figure}[b]
    \centering
    \includegraphics[width=1.0\linewidth]{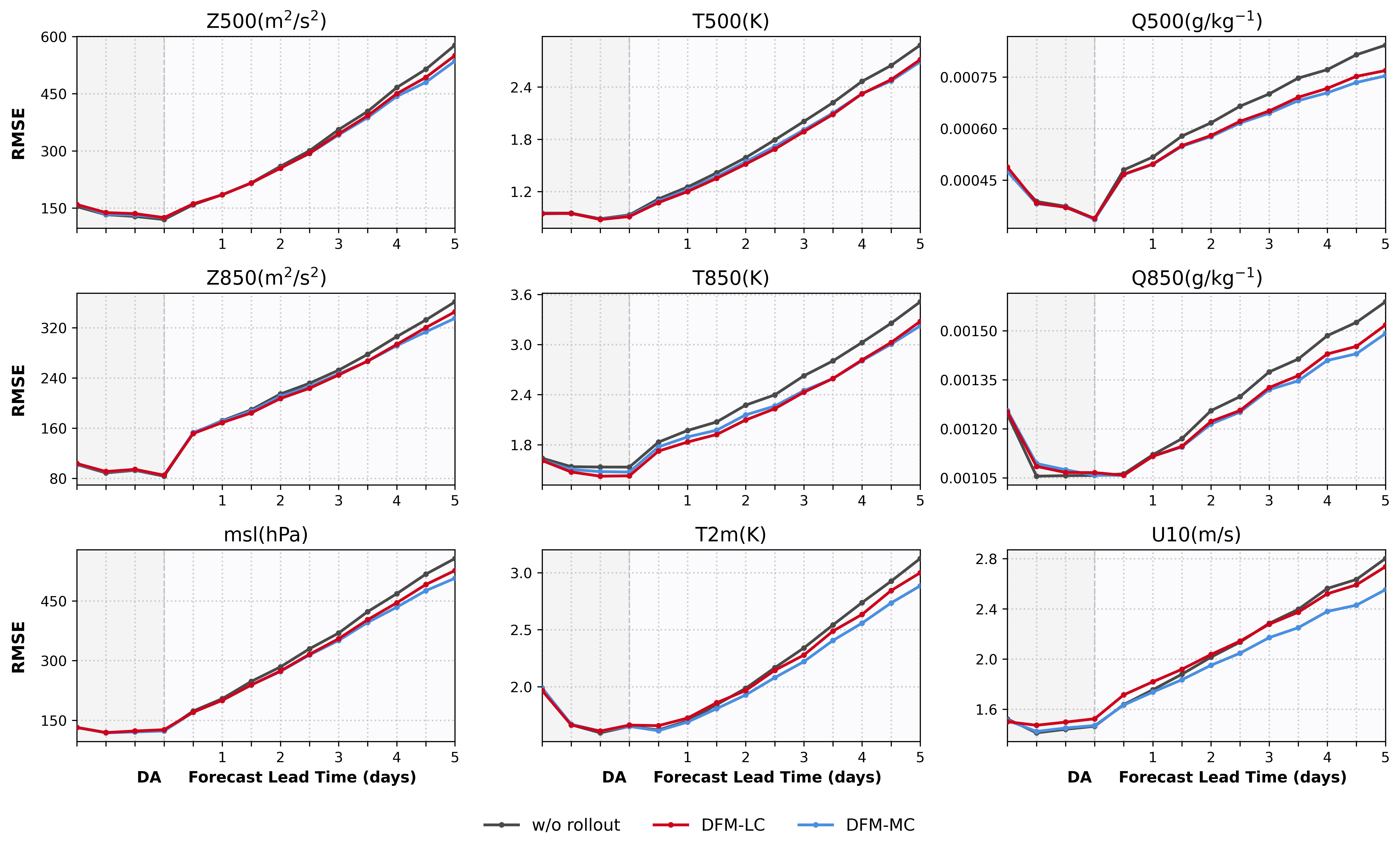}
    \caption{
    Analysis and forecast errors of cycling DA systems supported by different DFMs in real-observation experiments, using the 4DVar formulation in which only the model state is optimized. Observations consist of GDAS surface and radiosonde measurements. Analysis errors are evaluated against the 10\% of observations withheld from data assimilation, whereas forecast errors are evaluated against all observations.
    }
\end{figure}

\end{document}